%% file: main.tex
\documentclass[10pt,twocolumn,letterpaper]{article}

\usepackage[pagenumbers]{cvpr} 

\usepackage{graphicx}
\usepackage{amsmath}
\usepackage{amssymb}
\usepackage{booktabs}
\usepackage{subcaption}

\usepackage[table]{xcolor}

\usepackage[pagebackref,breaklinks,colorlinks]{hyperref}

\usepackage[capitalize]{cleveref}
\crefname{section}{Sec.}{Secs.}
\Crefname{section}{Section}{Sections}
\Crefname{table}{Table}{Tables}
\crefname{table}{Tab.}{Tabs.}

\newcommand{\edsr}{{EDSR-baseline}\xspace}

\newcommand{\subp}{{Sub-Pixel Conv.}\xspace}
\newcommand{\subpix}{{Sub-Pixel Convolution}\xspace}
\newcommand{\liif}{{LIIF}\xspace}
\newcommand{\lte}{{LTE}\xspace}
\newcommand{\metasr}{{Meta-SR}\xspace}

\newcommand{\cuf}{{CUF}\xspace}
\newcommand{\cufi}{{CUF-instantiated}\xspace}

\def\cvprPaperID{*****} 
\def\confName{CVPR}
\def\confYear{2022}

\input{preamble}
\begin{document}
\input{sec/0_metadata}
\maketitle

\input{sec/0_abstract}
\input{sec/1_introduction}

\input{sec/2_related}
\input{sec/3_method}

\input{sec/4_results}
\input{sec/5_conclusions}

{
    \small
    \bibliographystyle{ieee_fullname}
    \bibliography{macros,main}
}

\input{sec/X_supplementary}


\end{document}

%% file: preamble.tex
\usepackage{overpic}
\usepackage{enumitem} 
\usepackage{overpic} 
\usepackage{color}
\usepackage[skip=.5em]{caption} 
\usepackage{afterpage}
\usepackage{wrapfig}
\usepackage{placeins} 

\definecolor{turquoise}{cmyk}{0.65,0,0.1,0.3}
\definecolor{purple}{rgb}{0.65,0,0.65}
\definecolor{dark_green}{rgb}{0, 0.5, 0}
\definecolor{orange}{rgb}{0.8, 0.6, 0.2}
\definecolor{red}{rgb}{0.8, 0.2, 0.2}
\definecolor{darkred}{rgb}{0.6, 0.1, 0.05}
\definecolor{blueish}{rgb}{0.0, 0.3, .6}
\definecolor{greenish}{rgb}{0.0, .6, 0.3}
\definecolor{light_gray}{rgb}{0.7, 0.7, .7}
\definecolor{pink}{rgb}{1, 0, 1}
\definecolor{greyblue}{rgb}{0.25, 0.25, 1}
\definecolor{gold}{rgb}{0.7, 0.5, 0}

\newcommand{\CIRCLE}[1]{\raisebox{.5pt}{\footnotesize \textcircled{\raisebox{-.6pt}{#1}}}}

\DeclareMathOperator*{\argmin}{arg\,min}

\newcommand{\expect}{\mathbb{E}}
\newcommand{\real}{\mathbb{R}}

\usepackage{enumitem}
\setlist[itemize]{noitemsep,leftmargin=*,topsep=0in}
\setlist[enumerate]{noitemsep,leftmargin=*,topsep=0in}

\newcommand{\Figure}[1]{Figure~\ref{fig:#1}}

\newcommand{\Table}[1]{Table~\ref{tab:#1}}
\newcommand{\eq}[1]{(\ref{eq:#1})}

\newcommand{\Section}[1]{Section~\ref{sec:#1}}

\newcommand{\SupplementaryMaterial}{{\color{purple} appendix}\xspace}

\usepackage{blindtext}

\usepackage{lipsum}

\renewcommand{\paragraph}[1]{\vspace{1em}\noindent\textbf{#1}.}

\newcommand{\x}{\mathbf{x}}
\newcommand{\image}{\mathbf{I}}
\newcommand{\features}{\mathbf{F}}
\newcommand{\encoder}{\mathcal{E}}
\newcommand{\decoder}{\mathcal{D}}
\newcommand{\shuffle}{\mathcal{P}}
\newcommand{\upsampler}{\mathcal{U}}

\newcommand{\pars}{{\boldsymbol{\theta}}}
\newcommand{\channels}{C}
\newcommand{\kernel}{\mathcal{K}}
\newcommand{\relative}{\delta}
\newcommand{\posenc}{\Pi}
\newcommand{\offset}{k}
\newcommand{\scale}{s}
\newcommand{\kernelsize}{K}

%% file: sec/0_metadata.tex
\title{CUF: Continuous Upsampling Filters}


\author{Cristina Vasconcelos \quad Cengiz Oztireli  \quad Mark Matthews \\
Milad Hashemi \quad Kevin Swersky \quad Andrea Tagliasacchi
\\[1em]
Google Research
}

%% file: sec/0_abstract.tex
\begin{abstract}
Neural fields have rapidly been adopted for representing 3D signals, but their application to more classical 2D image-processing has been relatively limited.
In this paper, we consider one of the most important operations in image processing: upsampling.
In deep learning, learnable upsampling layers have extensively been used for single image super-resolution.
We propose to parameterize upsampling kernels as neural fields. This parameterization leads to a compact architecture that obtains a 40-fold reduction in the number of parameters when compared with competing arbitrary-scale super-resolution architectures. When upsampling images of size 256x256 we show that our architecture is 2x-10x more efficient than competing arbitrary-scale super-resolution architectures, and more efficient than sub-pixel convolutions when instantiated to a single-scale model.
In the general setting, these gains grow polynomially with the square of the target scale.
We validate our method on standard benchmarks showing such efficiency gains can be achieved without sacrifices in super-resolution performance.

\end{abstract}


%% file: sec/1_introduction.tex
\vspace{-2pt}
\section{Introduction}
\vspace{-2pt}
\label{sec:intro}

Neural-fields represent signals with coordinate-based neural-networks.
They have found application in a multitude of areas including 3D reconstruction\cite{mildenhall2020nerf}, novel-view synthesis~\cite{tancik2020fourfeat}, convolutions~\cite{Hermosilla_2018}, and many others~\cite{survey_xie2021neuralfields}.

\input{fig/teaser}

Recent research has investigated the use of neural fields in the context of \textit{single image super-resolution}~\cite{LIIF_chen2021,lte-lee2022}.\footnote{Thereon we assume \textit{single image} when talking about super-resolution.}
These models are based on multi-layer perceptrons conditioned on latent representation produced by encoders\footnote{These encoders were originally proposed for classical super-resolution applications, and include both convolutional~\cite{dong2014learning,EDSR_Lim_2017,RDN_zhang2020} as well as attentional ~\cite{ZhangRCAN2018,DaiSOCA2019,SWINIR_liang2021} architectures.}.
While such architectures allow for continuous-scale super-resolution, they require the execution of a conditional neural field for every pixel at the target resolution, making them unsuitable in applications with limited computational resources.
Further, such a large use of resources is not justified by a increase in performance compared to classical convolutional architectures such as sub-pixel convolutions~\cite{ShiCHTABRW16}.
In summary, neural fields have \textit{not yet} 
found widespread adoption as classical solutions are \CIRCLE{1} trivial to implement and \CIRCLE{2} more efficient. As they generally perform comparably, their usage is not justified in light of these points; see~\Figure{teaser}.

In this paper, we focus on overcoming these limitations, while noting that (regressive) super-resolution performance is in the saturation regime~(i.e. further improvements in image quality seem unlikely without relying on generative modeling~\cite{saharia2022image}, and small improvements in PSNR do not necessarily correlate with image quality).

\input{fig/continuous}
Our driving hypothesis is that super-resolution convolutional filters are highly correlated -- both spatially, as well as across scales.
Hence, representing such filters in the latent space of a conditional neural field can effectively capture and compress such correlations.
Given that neural fields encode continuous functions within neural networks, we call our filters \emph{Continuous Upsampling Filters} (CUFs).
While neural fields have proven effective in parameterizing~3D convolutions~\cite{Hermosilla_2018, WuQL19_PointConv, Boulch2020_ConvPoint},
we demonstrate that there are very significant savings to be had in the parameterization of~2D convolutions.

In implementing continuous upsampling filters, we draw inspiration from sub-pixel convolutions~\cite{ShiCHTABRW16}, and realize them via depth-wise convolutions.
This not only makes CUFs significantly more efficient than competing \textit{continuous} super-resolution architectures, but when instantiated to \textit{single-image super-resolution} they are, surprisingly, even more efficient than~``ad-hoc'' sub-pixel convolutions  with same number of input and output channels; see~\Figure{teaser}.

\paragraph{Contributions}
We investigate the use of neural fields as a parameterization of convolutional upsampling layers in super-resolution architectures, and show how:
\begin{itemize}
\item The continuity of neural-fields leads to training of compact convolutional architecture for continuous super-resolution~(similar super-resolution PSNR but with flop reduction that grows polynomially with the square of the target scale.).
\item Instantiating our continuous kernels into their discrete counterpart leads to an efficient inference pipeline, even more efficient in number of operations per target pixel than ad-hoc sub-pixel convolutions with same number of input and output channels.
\item Discrete cosine transforms can be used as an efficient replacement for Fourier bases in the implementation of positional encoding for neural fields.
\item These gains do not hinder reconstruction performance (PSNR), by carefully validating our method on standard benchmarks, and thoroughly ablating our design choices.
\end{itemize}

%% file: fig/teaser.tex
\begin{figure}[t]
\begin{center}
\includegraphics[width=\linewidth]{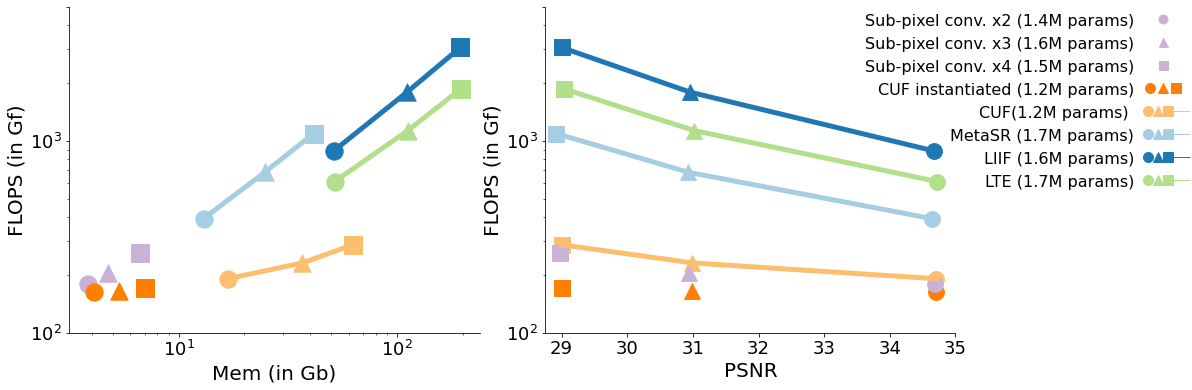}
\end{center}
\vspace{-1em}
\caption{
We report (left) memory and FLOPs for upsampling an $256\times256$ image by different scale factors ($2\times$,$3\times$,$4\times$) using integer scale upsamplers (\subpix ~\cite{ShiCHTABRW16} and \cufi) and arbitrary-scale upsamplers (\metasr~\cite{metasr_hu2019}, \liif~\cite{LIIF_chen2021}, \lte\cite{lte-lee2022} and \cuf), but the \textit{same} encoder backbone~(\edsr~\cite{EDSR_Lim_2017}); and 
(right) the relationship between each upsampler FLOPS and PSNR performance on DIV2k dataset.
Our arbitrary scale model is significantly lighter than other methods in the same class (i.e. continuous super-res).
Further, when instantiated for integer scale factors, our upsampler is even-more efficient than sub-pixel convolutions~\cite{ShiCHTABRW16}.
}
\label{fig:teaser}
\end{figure}

%% file: fig/continuous.tex
\begin{figure*}
\centering

{\setlength{\fboxsep}{0pt}
{ \setlength{\fboxrule}{1.2pt}
\begin{subfigure}{.9\textwidth}
\vfill
    \begin{subfigure}[ht]{0.033\textwidth}
        \includegraphics[width=\textwidth]{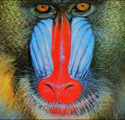}
        \caption*{$\times 1$}
    \end{subfigure}
    \begin{subfigure}[ht]{0.050\textwidth} 
        {\includegraphics[width=\textwidth]{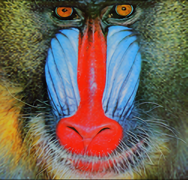}}
        \caption*{$\times 1.5$}
    \end{subfigure}
    \begin{subfigure}[ht]{0.067\textwidth} 
        \includegraphics[width=\textwidth]{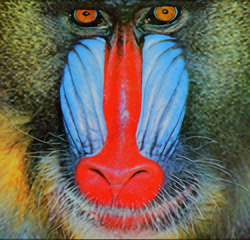}
        \caption*{${\color{blue}\times 2}$}
    \end{subfigure}
        \begin{subfigure}[ht]{0.083\textwidth}
        {\includegraphics[width=\textwidth]{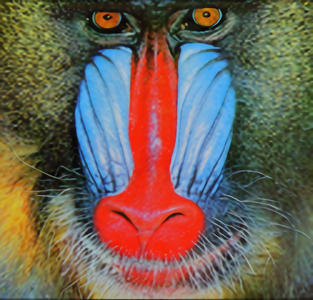}}
        \caption*{$\times 2.5$}
    \end{subfigure}
        \begin{subfigure}[ht]{0.10\textwidth}
        \includegraphics[width=\textwidth]{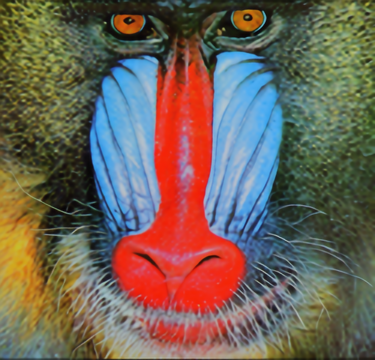}
        \caption*{${\color{blue}\times 3}$}
    \end{subfigure}
    \begin{subfigure}[ht]{0.117\textwidth}
        {\includegraphics[width=\textwidth]{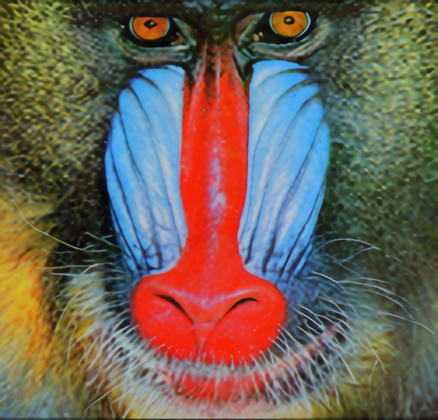}}
        \caption*{$\times$3.5}
    \end{subfigure}
        \begin{subfigure}[ht]{0.133\textwidth}
        \includegraphics[width=\textwidth]{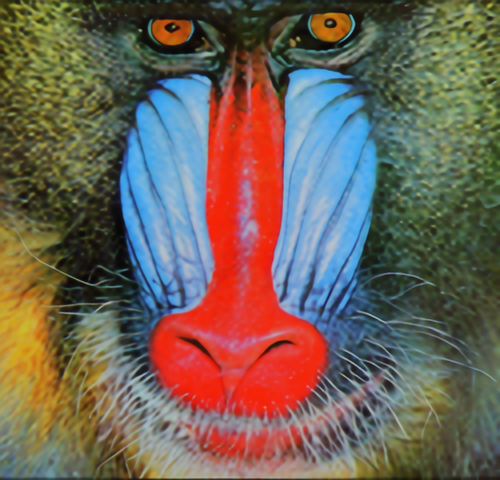}
        \caption*{${\color{blue} \times 4}$}
    \end{subfigure}
    \begin{subfigure}[ht]{0.15\textwidth}
        {\includegraphics[width=\textwidth]{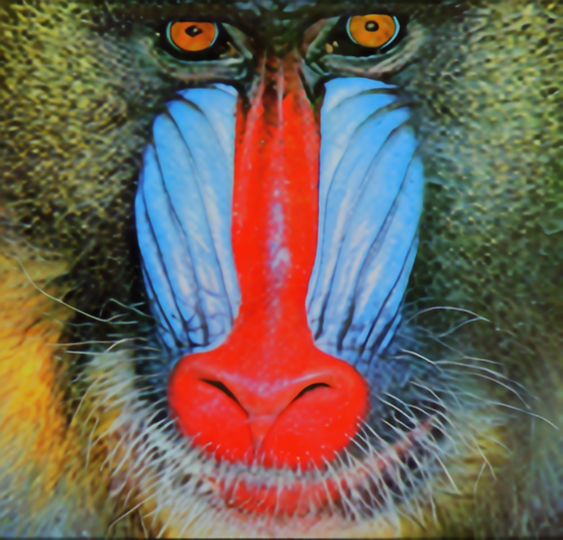}}
        \caption*{$\times$4.5}
    \end{subfigure}
    \begin{subfigure}[ht]{0.167\textwidth}
        {\includegraphics[width=\textwidth]{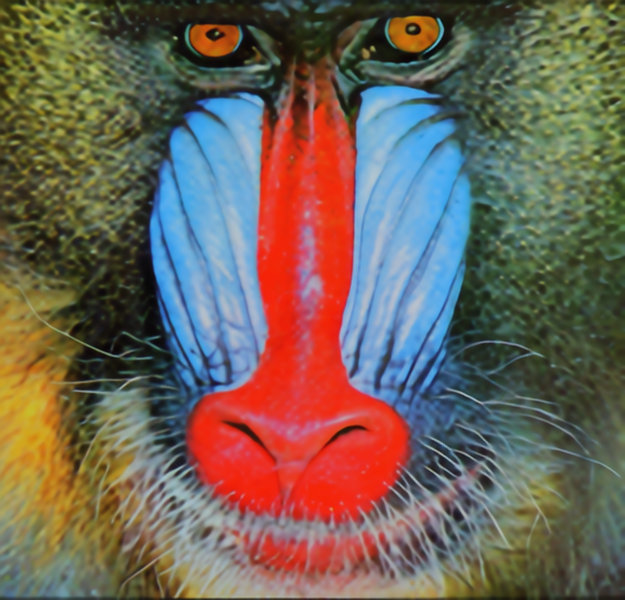}}
        \caption*{{\color{blue} $\times$5}}
    \end{subfigure}
\end{subfigure}
}}
\caption{
\textbf{Continuous super-resolution} -- 
Our architecture directly allows for an arbitrary and continuous choice of the super-resolution factor. Conversely, traditional methods perform this indirectly: they need to upsample to an {\color{blue}integer} factor, and then consequently downsample to the target resolution. More examples are available in the~\SupplementaryMaterial.
Note that $\times 4.5$ and $\times 5$ are upsampling factors not available in the training data.
}
\label{fig:continuous}
\end{figure*}

%% file: sec/2_related.tex
\vspace{-2pt}
\section{Related work}
\vspace{-2pt}
\label{sec:related}
A thorough coverage of single-image super-resolution can be found in the following surveys \cite{surveyWang2021, surveyAnwar2020, surveyBashir2021}; in what follows, we provide an overview of classical techniques for super-resolution based on deep learning.
As our model is based on neural-fields, we also point the reader to a survey in this topic~\cite{survey_xie2021neuralfields}, and below we discuss the existing works for super-resolution based on neural fields.

\paragraph{Single scale}
There are two main frameworks for super-res, which mainly differ in the placement of upsampling operators within the architecture.
\begin{itemize}
\item In the \textit{pre-upsampling framework}\cite{dong2014learning}, an initial upscaled image is obtained using a non-trainable upsampler (e.g. bicubic), which is then post-processed by a neural network. This enables arbitrary size/scaling, but can introduce side effects such as noise amplification and blurring.
\item In the \textit{post-upsampling framework}~\cite{ShiCHTABRW16}, an encoder that preserves the input spatial resolution is followed by a shallow upsampling component.
\end{itemize}
Note the computational complexity of pre-upsampling is significant, as these model operate directly at the target resolution. Consequently, post-upsampling is the de-facto mainstream approach~\cite{SRCNN_Dong2016, ESRGAN_Wang2018, EDSR_Lim_2017, RDN_zhang2020, ZhangRCAN2018, SWINIR_liang2021}.
For this reason, we investigate the impact of an implicit upsampler in \emph{post-upsampling frameworks}, and cover a diverse set of architectures ranging from large \cite{liu2021Swin}, to extremely lightweight models \cite{Du_2021_CVPR_abpn}.

\paragraph{Neural Fields}

\input{fig/prev_architectures}
\input{fig/architectures}

\input{tab/architecture_comparison}

The application of neural fields to super-resolution~\cite{metasr_hu2019} has enabled the design of architectures that support multiple target scales, as well as \textit{non-integer} super-resolution scales.
\begin{itemize}
    \item In \textit{meta super-resolution} \textbf{MetaSR}~\cite{metasr_hu2019}, a hyper-network is conditioned on the target scale factor and on the output pixel relative coordinates (ie. the difference between source and target grids covering the same normalized global space). Their model consists of: (1) transforming the input image using an encoder into deep features at the source resolution; (2) upsampling the deep features into the target space (using nearest-neighbors); (3) using the hyper-network to produce a filter set per target relative position and scale; (4) processing the enlarged target resolution feature map into the final output using the corresponding weights.

    \item In \textit{local implicit image function} \textbf{LIIF}~\cite{LIIF_chen2021}, the author eliminates the use of a hyper-network (MetaSR, step 3) by extending the sampled target resolution features (step 2) with extra channels representing the relative coordinates and target scale. Their combined features are further processed through a stack of five dense layers inspired by the multi-layer perceptron (MLP) architecture typically adopted as the end-to-end model in neural-fields formulations.
    
    \item In \textit{local texture estimator} \textbf{LTE}~\cite{lte-lee2022} the authors enhance the features fed as input to the MLP layers. In their model, the output of the encoder is processed by three extra trainable layers associated with the amplitude, frequency and phase of sin/cosine waves. The resulting projection in the target resolution layer is used as the input to an MLP that stacks $4$ fully-connected layers.
    In order to further improve the results they also adopt a global \textit{skip connection} with a bilinear up-scaled version of the input around the full model, such that the deep model focus on the computation of the residual between to the closed form approximation and the final result.
\end{itemize}
%
A visualization of these architectures can be found in~\Figure{prev_architectures}.
In practice, LIIF and LTE also increase the number of feature channels as their MLP layers have more neurons ($256$) than the number of channels produced as the encoder output ($64$ for encoders such as EDSR~\cite{EDSR_Lim_2017} and RDN~\cite{RDN_zhang2020}).
In contrast to these previous works, our upsampler:
\begin{itemize}
\item 
Moves the computational burden of arbitrary upsampling into the continuous upsampler operator while reducing
the operations performed in the target resolution space, preserving the number of channels produced by the deep encoder. 
\item 
Requires fewer operations than the single-scale sub-pixel convolution in the most typical application case of up-sampling an image by an integer scale factor, and producing results of comparable quality.
\item Adopts a neural-fields formulation that 
focuses the sensitivity of our hyper-network to fine-grained changes in the representation of scale and relative position (reducing spectral bias).
\end{itemize}




%% file: fig/prev_architectures.tex
\begin{figure*}[ht!]
\vspace{-1em}
\centering
\includegraphics[width=\linewidth]{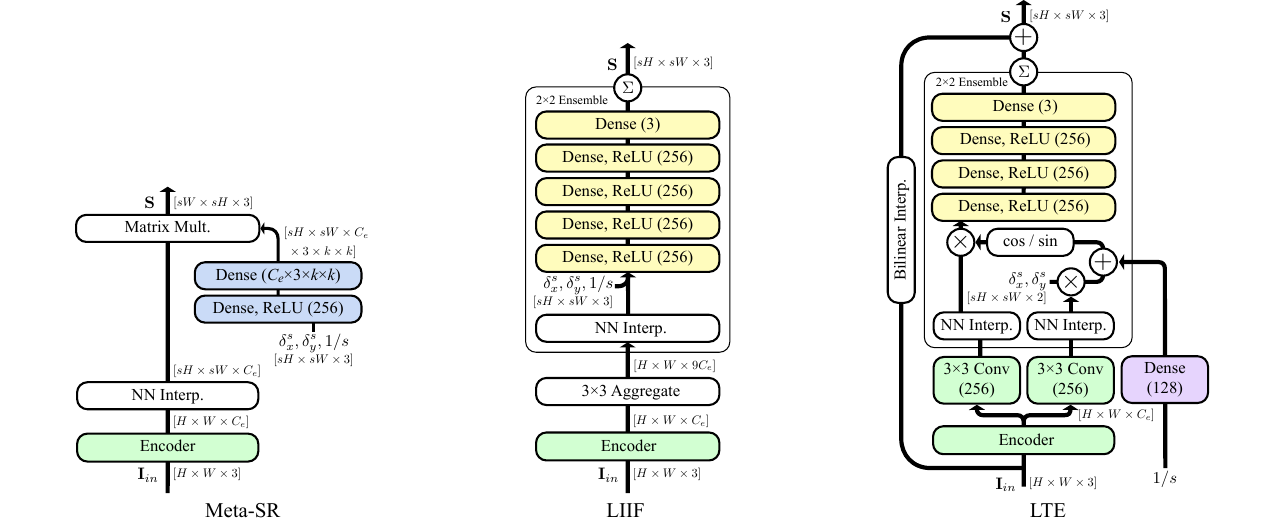}
\caption{
\textbf{Previous Architectures} -- 
A visual comparison of the super-resolution architectures of Meta-SR~\cite{metasr_hu2019}, local implicit image function LIIF~\cite{LIIF_chen2021}, and local texture estimators (LTE)~\cite{lte-lee2022}, shown in an isotropic scaling configuration. Components inside the  ``2$\times$2 Ensemble'' box are computed over the four closest pixels to compute a weighted average. The number of encoder channels, $C_e$, is dependent on the encoder chosen.
}
\label{fig:prev_architectures}
\vspace{-1em}
\end{figure*}

%% file: fig/architectures.tex
\begin{figure}[ht]
\centering
\includegraphics[width=\linewidth]{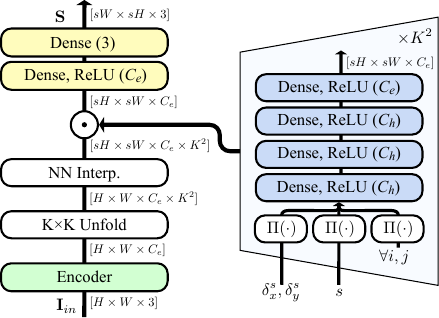}
\definecolor{learned_lr}{RGB}{211, 254, 211}
\definecolor{learned_sr}{RGB}{255, 252, 189}
\definecolor{learned_hyper}{RGB}{203, 219, 246}
\caption{
\textbf{Architecture} -- 
Continuous Upsampling Filters~(CUF) are based on a hyper-network.
For each pixel position in the high-res space, the hyper-network is computed $\kernelsize^2$ times, once for each position $i,j$ of the upsampling kernel (batched in practice). This can be reduced when upsampling with integer scales $s$, as is common in practice.
The Hyper-network output is aggregated into a tensor that is dot multiplied against the unfolded nearest-neighbor interpolation, computing a depth-wise convolution with per-pixel filters.
The number of encoder channels, $C_e$, is dependent on the encoder used. In our experiments, we set the number of hyper-network channels, $C_h = 32$.}
\label{fig:architectures}
\vspace{-1em}
\end{figure}

%% file: tab/architecture_comparison.tex
\begin{table*}
\centering
\resizebox{\linewidth}{!}{ 
\begin{tabular}{@{}l|cccc|cccc@{}}
&  \multicolumn{4}{c|}{Hyper Network} &
  \multicolumn{4}{|c}{Upsampler} \\
 & input & \#layers // \#neurons & output & params (K)
 & input & layers & neurons  & params (K)\\
 \hline
Meta-SR \cite{metasr_hu2019} & $[
\relative^\scale(\x),1/s]$ &  \cellcolor{orange!12}2 // 256 & $64\times9\times3$ & \cellcolor{orange!12}445 & 
$\encoder_\pars$ & convolution & --  & --\\
LIIF \cite{LIIF_chen2021} & -- &  -- & -- & -- & 
$[\encoder_\pars,\relative^\scale(\x),1/s]$ & 5 dense &  \cellcolor{red!12}256  & \cellcolor{orange!12}347
\\
LTE \cite{lte-lee2022} & -- &  -- & -- & -- & 
$h_a(\encoder_\pars)\{sin,cos\}(\relative^\scale(\x)\pi(h_b(\encoder_\pars)+h_c(1/s))$ & 4 dense &  \cellcolor{red!12} 256 & \cellcolor{red!12}494
\\
\hline
\textbf{Ours}:

& $[\posenc(\relative^\scale(\x)),\posenc(\scale), \posenc(\offset)]$
& \cellcolor{green!12} 4 // 32& 64 &
\cellcolor{green!12}6
& $\encoder_\pars$ & d-w conv. + 2 dense  & \cellcolor{green!12}64 &\cellcolor{green!12}4
\\
%
\end{tabular}
} 
\caption{
\textbf{Super-resolution with neural-fields:}
An overview of the main characteristics of neural field based architectures. 
Square brackets represent concatenation.
In LTE~\cite{lte-lee2022}, $h_a$,$h_b$ are $3\times3$ convolutional layers and $h_c$ is a linear layer.
%
} 
\label{tab:archtectures}
\vspace{-1em}
\end{table*}

%% file: sec/3_method.tex
\vspace{-2pt}
\section{Method}
\vspace{-2pt}
Given a target scale $\scale$, we aim to produce an upscaled image of size~$\scale H {\times} \scale W {\times} 3$ given an input image of size~$H {\times} W {\times} 3$.
Within the context of super-resolution, our core contribution is the introduction of a novel \textit{learnable upsampling} layer.
The upsampling layer can be interpreted as a decoder in a classical encoder-decoder architecture.
We briefly review our encoder architecture in \Section{encoder}, and detail our decoder in \Section{decoder}.
Note that our analysis focuses on the \textit{decoder}, as a variety of encoders can be used.
In \Section{analysis}, we then perform a conceptual comparison between our architecture, shown in \Figure{architectures}, and others based on neural fields, which are visually summarized in~\Figure{prev_architectures}. 

\paragraph{Training}
Our network with parameters $\pars$ is trained to \textit{regress} a high resolution image $\tilde\image^\scale$ matching the ground truth 
$\tilde\image^\scale_\text{gt}$ at random positions $\x$ given the low resolution input image $\image$:
\begin{equation}
\argmin_\pars \:\:
\expect_\image \:
\expect_\scale \:
\expect_\x \:
\| \tilde\image^\scale(\x; \image, \pars) - \tilde\image^\scale_\text{gt}(\x) \|_2^1
\end{equation}

\subsection{Encoder}
\label{sec:encoder}
Towards this objective, we first process our input image with an encoder producing $\channels$-dimensional features:
\begin{equation}
\encoder_\pars : \real^{H {\times} W {\times} 3} \rightarrow \real^{H {\times} W {\times} C}
\end{equation}
and then unfold a $k {\times} k$ spatial neighborhood of $C$-dimensional features into a tensor of 
$\channels {\times} k^2$ channels. Finally, we upsample the encoded image using nearest-neighbour interpolation. We define this step as:
\begin{equation}
\upsampler : \real^{H {\times} W {\times} \channels} \rightarrow \real^{sH {\times} sW {\times} \channels \times k^2}
\end{equation}
The unfolding part of this procedure leads to \textit{feature maps} that will allow us to implement depth-wise, spatial convolutions as dot products, as is typically done for low-level neural network implementations. We define this whole feature extraction procedure as:
\begin{equation}
\features = \upsampler(\encoder_\pars(\image))
\end{equation}
 

\input{fig/continuous_filters}
\subsection{Decoder -- ``Continuous Upsampling Filters''}
\label{sec:decoder}
We draw our inspiration from classical sub-pixel convolutions~\cite{ShiCHTABRW16}, and combine it with recent research that apply neural fields for continuous super-resolution~\cite{LIIF_chen2021, lte-lee2022}.
Differently from these architectures, we achieve this objective by instantiating upsampling kernels $\kernel$ via \textit{hyper-networks}~\cite{ha2016hypernetworks}.
Our super-resolution network is algebraically expressed as:
\begin{equation}
\tilde\image^\scale(\x; \image, \pars) = \decoder_\pars( 
\kernel_\pars(\relative^\scale(\x); \scale)
\cdot
\features(\lfloor \x / \scale \rfloor))
\end{equation}
where $\x$ indexes pixel locations in the target resolution, and $\relative^\scale(\x) {=} \bmod(\x, \scale) / \scale$ renders our convolutional filters translation invariant, and the dot product implements a spatial convolution.
The network $\decoder$ is a point-wise layer that maps~(super-resolved) features back to RGB values:

\begin{equation}
\decoder_\pars :
\real^{C}
\rightarrow 
\real^{3}
\label{eq:pointwise}
\end{equation}

This is applied to a grid of coordinates of size $\real^{\scale H {\times} \scale W {\times} C}$ to get an image of size $\real^{\scale H {\times} \scale W {\times} 3}$.
The coordinate $\x$ is \textit{continuous}, hence $\kernel_\pars$ is a \textit{neural field} parameterization of a convolutional kernel mapping~(continuous) spatial \textit{offsets} and~(continuous) \textit{scales} to convolutional \textit{weights}:
\begin{equation}
\kernel_\pars : [0,1]^2 \times \real^+ \rightarrow \real^{\channels \times \kernelsize^2}
\label{eq:cups}
\end{equation}


\input{tab/benchmarks}
\input{tab/sota}

\paragraph{Continuous kernel indexing} 
We can take our continuous formulation a step further by introducing a continuous parametrization of its kernel indexes $k_i, k_j$ indexing our $\kernelsize \times \kernelsize$ convolution weight entries.
Thus, the hyper-network representing the convolution field becomes:
\begin{equation}
\hat\kernel_\pars : [0,1]^2
\times \{0,\ldots,\kernelsize-1\}^2 
\times \real^+
\to \mathbb{R}^C
\label{eq:continuouskernel}
\end{equation}
and $\kernel_\pars$ can then be constructed by $9$ invocations to $\hat\kernel_\pars$ in the case of a $\kernelsize{=}3$, and $\kernelsize^2$ invocations in the general setting. We use $\kernelsize{=}3$ throughout the paper in order to facilitate a direct comparison to pre-existing baseline models. 

Equation \eqref{eq:cups} and Equation \eqref{eq:continuouskernel} are both valid ways of generating spatial kernels. The first can be seen as a multi-headed hypernetwork, while the second uses input conditioning to generate the kernel values. We use the latter in our experiments, as the layers of nonlinearities provide additional expressiveness compared to the linear transformation used in the multi-headed version. A comparison between the two is presented in \autoref{sup:kin_kout}.


\subsection{Positional encoding}
\label{sec:posenc}
Naively representing signals with MLPs leads to ``spectral bias''~\cite{pmlr-v97-rahaman19a}: the overall difficulty these networks have in representing high-frequency signals.
Hence we implement $\kernel_\pars$~as:
\begin{equation}
\kernel_\pars(\relative^\scale(\x), s, \offset) = \text{MLP}( \posenc(\relative^\scale(\x)),\posenc(\scale),\posenc(\offset) )
\end{equation}
where we apply positional encoding $\posenc(\cdot)$ to the MLP inputs.
Several variants of positional encoding exist, including Fourier~\cite{mildenhall2020nerf} and random Fourier~\cite{tancik2020fourfeat} variants.
However, as we strive for efficiency, we take inspiration from classical signal processing (e.g. JPEG) and instead employ a cosine-only transformation.
Given a scalar $f_n$ sampled uniformly in the range $[0, f_\text{max}]$, a scalar quantity $z \in [0,1]$ is encoded as the~(sorted) vector:
\begin{equation}
\posenc(z) =
\left\{
cos\left( \frac{(2z+1)f_n\pi}{2} \right)
\right\}_{n=1}^N
\end{equation}
By eliminating the imaginary component from the Fourier basis, we show in~\Section{posenc_results} how this reduces the number of trainable parameters of the first hyper-network layer by half, without affecting super-resolution quality.


\subsection{Analysis}
\label{sec:analysis}
We now perform a conceptual comparison of our network architecture to the commonly used \subpix, as well as several super-resolution approaches based on neural-fields. We further note how continuous upsampling filters can be instantiated to generate filters compatible with \subpix.

\paragraph{Comparison to \subpix~\cite{ShiCHTABRW16}}
\subpix is the most commonly used super-resolution operator, especially when efficiency is critical.
This is achieved by initially employing an expansion convolution 
that generates a feature map with $s^2N_{out}$ channels, where $N_{out}$ is a hyper parameter defining the target number of channels of the \subpix operation.
%
\begin{figure} 
\begin{center}
\includegraphics[width=0.5\columnwidth]{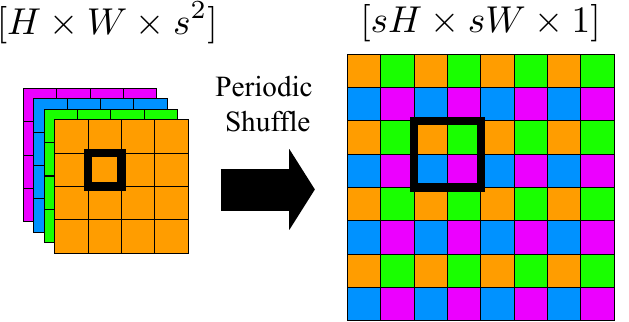} 
\end{center}
\vspace{-1em}
\caption{A simple illustration of periodic shuffling used in \subpix.}
\label{fig:subpix}
\vspace{-1em}
\end{figure}
Next, the periodic shuffling operator~\footnote{
also called \emph{pixel-shuffle}~(PyTorch) or depth-to-space~(TensorFlow).} re-arranges the channels to produce a higher-resolution feature map (see \Figure{subpix}):
\begin{equation}
\shuffle : \real^{H {\times} W {\times} \channels} \rightarrow \real^{\scale H {\times} \scale W {\times} \channels / \scale^2}
\end{equation}
Similar to \eq{pointwise}, whenever  $N_{out}$ is chosen as different from the number of output colors, the resulting feature map is further projected by point-wise convolutions. Often, $N_{out}$ is taken as the same as the number of channels as produced by the encoder $\encoder$ for architectures adopting \subpix and targeting high quality results~\cite{EDSR_Lim_2017,RDN_zhang2020,SWINIR_liang2021}.
We note that the \subpix design based on the expansion convolution does not enforce spatial correlations between neighboring pixels (here we refer to neighborhood in the output, high resolution, domain) and need to be learned from training data.
Conversely, these correlations are inherently captured by our continuous convolutional filters.
\quad

Further, in the integer scale setup where hyper-networks can be pre-instantiated, CUFs are computationally \textit{more} efficient than sub-pixel convolutions whenever $\kernelsize^2 + N_{in} + N_{out} < N_{out}\times \kernelsize^2$ (in the worst case scenario) as the cost associated with pre-computing the weights for a given integer up-sampling scale can be neglected when compared to the operations performed on the image grid (see \autoref{sup:ops}).

\paragraph{Comparisons to LIIF~\cite{LIIF_chen2021} and LTE~\cite{lte-lee2022}}
In comparison to these works, we make the main network shallower ($1$ depth-wise continuous convolution and $2$ dense layers) in order to reduce the number of layers that operate at the target spatial resolution and at the encoder's channel resolution.
At the same time, our upsampler head is \textit{considerably} lighter than previous arbitrary scale methods, not only by the use of a depth-wise convolution but also by keeping the same number of channels produced by the features encoder in the main network ($C_e=64$, which results in dense layers $16\times$ cheaper than its LIIF and LTE counterparts using $256$ neurons).
When performing \textit{non-integer} upsampling, the costs with processing the hyper-network layers are proportional to the target image resolution, but still smaller than a single dense layer of LIFF and LTE heads, due to the adoption of a reduced number of channels (as $C_h=32$, each of CUF's hypernetwork dense layers is $64\times$ cheaper than LIIF and LTE layers using $256$ neurons).

\paragraph{Instantiating CUFs} 
At inference time, when targeting an \textit{integer} upscaling factor $\scale$, the hyper-network representing $\kernel$
can be \textit{queried}  to retrieve the weights corresponding to $s^2$ relative subpixel positions
as an initialization step during pre-processing. 
The retrieved weights are re-used across all pixels taking advantage of the existent periodicity. 
Thus, in the \cufi architecture the continuous kernel is replaced at test time with a discrete depth-wise convolution, followed by a pixel shuffling operation, in contrast with the unfolding operator used in the regular, fully continuous CUF described in this paper. Otherwise, the architecture remains the same.
The costs associated with the hyper network at initialization can be neglected as $s^2<<sH\times sW$. 
In this setting, our model becomes as efficient as a sub-pixel-convolution architecture, while retaining the aforementioned continuous modeling properties at training time.


%% file: fig/continuous_filters.tex
\begin{figure}[t]
\centering
\begin{subfigure}{0.5\textwidth}
\vfill
    \begin{subfigure}[ht]{0.32\textwidth}
        \includegraphics[width=\textwidth]{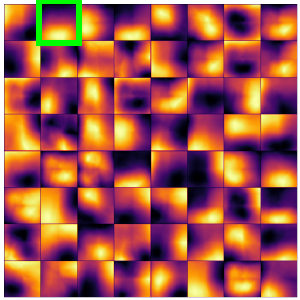}
        \caption*{all filters}
    \end{subfigure}
    \begin{subfigure}[ht]{0.32\textwidth}
        \includegraphics[width=\textwidth]{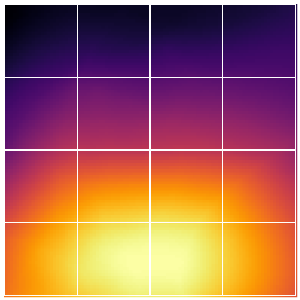}
        \caption*{filter \#2 - continuous}
    \end{subfigure}
        \begin{subfigure}[ht]{0.32\textwidth}
        \includegraphics[width=\textwidth]{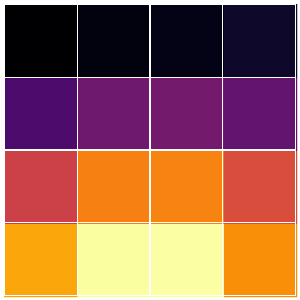}
        \caption*{filter \#2 - discretized}
    \end{subfigure}
\end{subfigure}
\caption{
\textbf{Continuous filters} --
CUF filters are continuous functions stored in neural fields. We visualize our filters by densely sampling each on the $[0,1]^2$ domain to highlight their continuous nature (left). Close-up of an exemplar filter (middle). 
Filters are discretized by sampling at pixel centers (right), allowing them to be used in classical fixed-scale super-res architectures.
}

\label{fig:continuous_filters}
\vspace{-1em}
\end{figure}

%% file: tab/benchmarks.tex
\newcommand{\firstrank}{{\cellcolor{green!30}}}
\newcommand{\secondrank}{{\cellcolor{green!8}}}
\begin{table*}[ht]
\centering
\resizebox{\textwidth}{!}{
\footnotesize
\begin{tabular}{lll||c|c|c|c|c|c|c|c|c|c|c|c|c|c|c|c}
  & & &
  \multicolumn{4}{c|}{Set5} &
  \multicolumn{4}{c|}{Set14} &
  \multicolumn{4}{c|}{BSD100} & 
  \multicolumn{4}{c}{Urban100}  \\
 Encoder & Upsampler & Self Ens. & \tiny{$\times$ 2} &   \tiny{$\times$ 3} &   \tiny{$\times$ 4} &  \tiny{$\times$ 6} &  
  \tiny{$\times$ 2} &   \tiny{$\times$ 3} &   \tiny{$\times$ 4} & \tiny{$\times$ 6} &   
    \tiny{$\times$ 2} &   \tiny{$\times$ 3} &   \tiny{$\times$ 4} & \tiny{$\times$ 6} &    
      \tiny{$\times$ 2} &   \tiny{$\times$ 3} &   \tiny{$\times$ 4} & \tiny{$\times$ 6}
      \\
 \hline
RDN & fixed-scale \cite{RDN_zhang2020}&  &  
\firstrank 38.24& \secondrank 34.71&\secondrank 32.47 &-&
 \firstrank 34.01&
 \secondrank 30.57&
 \secondrank 28.81&-&
 \secondrank 32.34&
 \secondrank 29.26&
 \secondrank 27.72&-&
 32.89&28.80
&\secondrank 26.61&-
\\

& MetaSR \cite{metasr_hu2019}&  & 
38.22&
34.63&
32.38&
\secondrank 29.04&
33.98&
30.54&
28.78&
\secondrank 26.51&
32.33&
\secondrank 29.26&
27.71&
\secondrank 25.90&
\secondrank 32.92&
\secondrank 28.82&
26.55&
\secondrank 23.99
\\
&CUF (ours) &  
& \secondrank 38.23 
& \firstrank 34.72 
& \firstrank 32.54 
& \firstrank 29.25 
& \secondrank 33.99
& \firstrank 30.58
& \firstrank 28.86
& \firstrank 26.70
& \firstrank 32.35
& \firstrank 29.29
& \firstrank 27.76
& \firstrank 25.99
& \firstrank 33.01
& \firstrank 28.91
& \firstrank 26.75
& \firstrank 24.23
\\
\hline
 & fixed-scale \cite{RDN_zhang2020}& +$_\text{geo}$&  
\firstrank 38.30 &
\secondrank 
\secondrank 34.78 &\secondrank 32.61&-&
\firstrank 34.10&
\firstrank 30.67&
\firstrank 28.92&-&
\firstrank 32.40&
\firstrank 29.33&
27.75&-&
 \secondrank 33.09&
  \secondrank 29.00&
   \secondrank 26.82&-
\\
& LIIF \cite{LIIF_chen2021} & +$_\text{loc}$& 
38.17&34.68&32.50&29.15&
33.97&30.53&28.80&26.64&
32.32&29.26&27.74&25.98&
32.87&28.82&26.68&24.20
\\
& LTE \cite{lte-lee2022}& +$_\text{loc}$&
38.23&34.72&
\secondrank 
32.61&\firstrank 29.32&
\secondrank 34.09&30.58&28.88&
\secondrank 26.71&
32.36&
\secondrank 29.30&
\secondrank 27.77&
 \secondrank 26.01&
33.04&28.97&26.81&
\secondrank 24.28
\\ 
& CUF (ours)& +$_\text{geo}$ &
\secondrank 38.28 &
\firstrank  34.80 &
\firstrank 32.63 &
\secondrank 29.27 &
\secondrank 34.08 &
\secondrank 30.65 &
\firstrank 28.92 &
\firstrank 26.74 &
\secondrank 32.39 &
\firstrank 29.33 &
\firstrank 27.80 &
\firstrank 26.03 &
\firstrank 33.16 &
\firstrank 29.05 &
\firstrank 26.87 &
\firstrank 24.32 
\\
 \hline
SwinIR & fixed-scale~\cite{ShiCHTABRW16}  &
& \firstrank  38.35 
&\firstrank 34.89 
&\secondrank 32.72 
&-
& \secondrank 34.14 
& \secondrank 30.77 
& \secondrank 28.94
&-
& \secondrank 32.44 
& \secondrank 29.37 
& \secondrank 27.83
&-
& \secondrank 33.40 
& \secondrank 29.29 
& \secondrank 27.07
&-
\\

&MetaSR \cite{metasr_hu2019}& &
38.26&34.77&32.47&\secondrank 29.09&
\secondrank 34.14&30.66&28.85&\secondrank 26.58&
32.39&29.31&27.75&\secondrank 25.94&
33.29&29.12&26.76&\secondrank 24.16
\\ 
& CUF (ours) &
&\secondrank 38.34 &\secondrank 34.88 &
\firstrank 32.80 &
\firstrank 29.53 &
\firstrank 34.29 &
\firstrank 30.79 &
\firstrank 29.02 &
\firstrank 26.85 & 
\firstrank 32.45 &
\firstrank 29.38 &
\firstrank 27.85 &
\firstrank 26.09 &
\firstrank 33.54 &
\firstrank 29.45 &
\firstrank 27.24 &
\firstrank 24.62 
\\
\hline
& fixed-scale~\cite{ShiCHTABRW16}  &+$_\text{geo}$
& \firstrank 38.38   
& \firstrank 34.95 
& \secondrank 32.81 
&-
& 34.24  
& \secondrank 30.83 
& 29.02 
&-
& \firstrank  32.47  
&  \firstrank  29.41 
&  \secondrank  27.87 
&-
& \secondrank 33.51  
& \secondrank 29.42 
&27.21 
&-
\\

&LIIF\cite{LIIF_chen2021} &+$_\text{loc}$& 
38.28&34.87&32.73&29.46&
34.14&30.75&28.98&26.82&
32.39&29.34&27.84&26.07&
33.36&29.33&27.15&24.59
\\
&LTE+lc \cite{lte-lee2022}&+$_\text{loc}$&
\secondrank 38.33&
34.89&
\secondrank 32.81&
\secondrank 29.50&
\secondrank 34.25&30.80&
\firstrank 29.06&
\secondrank 26.86&
\secondrank 32.44&29.39&27.86&
\secondrank 26.09&
33.50&
29.41&
\secondrank 27.24&
\secondrank 24.62
\\ 
& CUF (ours)+ &+$_\text{geo}$ &
\firstrank 38.38 &
\secondrank 34.92 &
\firstrank 32.83 &
\firstrank 29.57 &
\firstrank 34.33 &
\firstrank 30.84 &
\secondrank 29.05 &
\firstrank 26.91 &
\firstrank 32.47 &
\secondrank 29.40 &
\firstrank 27.88 &
\firstrank 26.11 &
\firstrank 33.65 &
\firstrank 29.55 &
\firstrank 27.32 &
\firstrank 24.69 
 \\
\end{tabular}}
\caption{
\textbf{Out-of-domain evaluation} -- performance on datasets not seen during training.
Metrics computed on the luminance channel in the YCbCr color space, following previous work.
Models using self-ensemble are marked with $+_{geo}$ and models using local self-ensemble are marked with $+_{loc}$. 
\textbf{Legend:} \colorbox{green!30}{best} \colorbox{green!6}{2\textsuperscript{nd} best} 
}
\label{tab:benchmarks}
\vspace{-1em}
\end{table*} 

%% file: tab/sota.tex
\newcommand{\first}{{\cellcolor{green!30}}}
\newcommand{\second}{{\cellcolor{green!8}}}

\begin{table}[ht!]
\centering
\resizebox{\linewidth}{!}{
\begin{tabular}{lll||c|c|c|c|c|c} 
\hline
 \multicolumn{8}{c}{Multi-scale up-sampling methods - DIV2k} \\
 \hline
 Encoder & Upsampler & Ens. &\multicolumn{3}{c|}{seen scales}& \multicolumn{3}{c}{unseen scales} \\
 & & & \tiny{$\times$ 2} &   \tiny{$\times$ 3} &   \tiny{$\times$ 4} &  \tiny{$\times$ 6} &  
 \tiny{$\times$ 12} & \tiny{$\times$ 18} 
\\
  \hline 
 Bicubic
 & & &
 31.01&
 28.22&
 26.66&
 24.82&
 22.27&
 21.00
  \\
\hline 
EDSR-b. 
& Sub-pixel conv.
& & \second 34.69 & \second 30.94 & \second 28.97 & -&--&--
\\ 
& MetaSR  
& &34.64& 30.93& 28.92& \second 26.61& \second 23.55& \second 22.03
\\
& CUF (ours) 
&
&\first 34.70
&\first 30.99
&\first 29.01
&\first 26.76
&\first 23.73
&\first 22.20
\\
\hline
& Sub-pixel conv.
&
+$_\text{geo}$ &
 \second 34.78 &\second  31.03& \second 29.06&--&--&--
 \\
& LIIF
&  +$_\text{loc}$
& 34.67& 30.96& 29.00& 26.75& \second 23.71& 22.17
\\
& LTE
&  +$_\text{loc}$
 & 34.72& 31.02 & 29.04 & 
 \second 26.81 &
 \first 23.78 & \second 22.23 
 \\  

& CUF (ours) & +$_\text{geo}$
&\first 34.79 & \first 31.07 & \first 29.09 &\first 26.82 &\first 23.78 &\first 22.24 
\\
\hline
  RDN
  &
  Sub-pixel conv.
  & &
\second 35.01&31.22 & 29.20& --& --& --
\\
& MetaSR 
&  &
 35.00&
 \second 31.27&
 \second 29.25&
 \second 26.88&
 \second 23.73&
 \second 22.18
\\
& CUF (ours) & &
\first 35.03&
\first 31.31&
\first 29.32&
\first 27.03&
\first 23.94&
\first 22.38
\\
  \hline
  & Sub-pixel conv.
  & +$_\text{geo}$
&\second 35.10
&\second 31.33
&\second 29.31 &--&--&--
\\

& LIIF 
&  +$_\text{loc}$
&  
 34.99&
 31.26&
 29.27&
 26.99&
 23.89&
 22.34
 \\
 & LTE 
 &  +$_\text{loc}$
 & 35.04&
 31.32&
 29.33&
 \second 27.04&
 \second 23.95&
 \second 22.40
 \\ 
& CUF (ours)&+$_\text{geo}$& 
\first 35.11 &
\first 31.39 &
\first 29.39 &
\first 27.09 &
\first 23.99 &
\first 22.42 
\\
\hline
  
SwinIR
&
Sub-pixel conv. 
& &
\first 35.28 & \second 31.47 & \second 29.40 & --& --& --
\\
 
& MetaSR 
& &
35.15&
 31.40&
 29.33&
 26.94&
 23.80&
 22.26
\\
& CUF (ours) & &
\second  35.26 &
\first 31.52 &
\first 29.52&
\first 27.19 &
\first 24.07 &
\first 22.49 
 \\
\hline
& Sub-pixel conv.
& +$_\text{geo}$
&\first 35.33 
&\second 31.52 
&29.44 &--&--&--
\\

& LIIF 
&  +$_\text{loc}$
& 35.17&
 31.46&
 29.46&
 27.15&
 24.02&
 22.43
 \\
 & LTE 
 &  +$_\text{loc}$
 & 35.24&
 31.50&
 \second 29.51&
 \second 27.20&
 \second 24.09&
 \second 22.50
 \\ 

 & CUF (ours)& +$_\text{geo}$&
\second 35.31& 
\first 31.56&
\first 29.56&
\first 27.23&
\first 24.10 &
\first 22.52 
 \\
 \end{tabular}
}
\caption{
\textbf{In domain evaluation:} tests on DIV2K's validation subset~\cite{DIV2k_Timofte2017}
on scales seen (2$\times$ -- 4$\times$) and unseen during training (6$\times$ -- 30$\times$) .
Metrics taken on the RGB space. Results taken using geometric self-ensemble \cite{EDSR_Lim_2017} are marked with '$+_\text{geo}$' and local-ensemble \cite{LIIF_chen2021} with '$+_\text{loc}$'. 
\textbf{Legend:} \colorbox{green!30}{best} \colorbox{green!6}{2\textsuperscript{nd} best} 
}
\label{tab:sota}
\vspace{-1em}
\end{table}

%% file: sec/4_results.tex
\vspace{-2pt}
\section{Results}
\vspace{-2pt}
In this section we describe our experimental setup in terms of backbone encoders and training/validation dataset, perform careful comparisons to the state-of-the-art (\Section{sota}), and additional evaluation towards the implementation of lightweight super-res architectures~(\Section{lightweight-architectures}). We conclude by performing a thorough ablation~(\Section{ablations}).


\input{tab/params}
\input{fig/qualitative.tex}

\paragraph{Encoders}
We apply CUFs to a variety of \textit{encoders}, to both show its generality, as well as its performance gains in a variety of settings.
\begin{itemize}
\item A state-of-the-art encoder for super-res named SwinIR   \cite{SWINIR_liang2021} based on Swin-transformers \cite{liu2021Swin}.
\item Well-known convolution-based encoders, namely EDSR-baseline~\cite{EDSR_Lim_2017} and Residual Dense Networks (RDN)~\cite{RDN_zhang2020}. 
\item Architectures for lightweight and extremely lightweight inference, respectively SwinT-lightweight~\cite{SWINIR_liang2021} and ABPN~\cite{Du_2021_CVPR_abpn}.
\end{itemize}
Table \ref{tab:params} contrasts the ablated encoders' composition and size. Hyperparameter settings can be found in the \SupplementaryMaterial.


\paragraph{Datasets}
All models are trained using the DIV2K dataset training subset ($800$k images), introduced at the~NTIRE~2017 Super Resolution Challenge~\cite{DIV2k_Timofte2017}.
Similarly to previous works, each image is randomly cropped 20 times per epoch and augmented with random flips, and 90 degrees rotations. 
We report peak signal-to-noise ratio~(PSNR) results on the DIV2K validation set (100 images)\footnote{The models we compare to also do not report results on the test-set, which is no longer publicly available.}, in the \emph{RGB} space.
To understand the generalizability of our experimental results, we test our models on additional well known datasets (Set5 \cite{SET5_Bevilacqua2012} -- 5 images;  
Set14 \cite{Set14_Zeyde2010} -- 14 images; 
B100 \cite{B100_Martin2001}  -- 100 images;
and Urban100 \cite{Urban100_Huang2015} -- 100 images), in which, following previous works, PSNR is measured in the \emph{YCbCr} luminance channel.


\paragraph{Ensembling}
Whenever demarked by `$+_\text{geo}$', results incorporate the (commonly adopted) \textit{Geometric Self-ensemble}~\cite{EDSR_Lim_2017}, where the results of rotated versions of the input images are averaged at test time. LIIF and LTE results adopt \textit{Local Self-ensemble} (`$+_\text{loc}$'), where the result of applying the upsampler at $4$ shifted grid points is averaged, resulting in a $\times 4$ increase in computational complexity at training/test time.
It may also introduce confounding factors to the optimization process that invalidate direct comparison to the fixed-scale upsamplers, as during back-propagation the encoder gradients are averaged. 
In~\autoref{sup:ensemble} we combine these models.


\subsection{State-of-the-art comparisons}
\label{sec:sota}
\Table{benchmarks} and \Table{sota} show the quantitative comparisons, while qualitative results can be observed in \Figure{continuous}, \Figure{qualitative} and \autoref{sup:qualitative}.
\Table{sota} contains \textit{in-domain} results: models trained on DIV2k and tested on DIV2k (train/test split), while \Table{benchmarks} presents out-of-domain results: models are trained on DIV2k, but tested on different datasets.
\quad 
Note that upsampling ratios larger than $\times 4$ are not provided at training time. The models adopting sub-pixel convolutions are fixed-scale models, that is, one model is trained per target scale and thus do not generalize to unseen or fractional scales. On the remaining rows we present results of arbitrary-scale upsamplers (MetaSR~\cite{metasr_hu2019}, LIIF~\cite{LIIF_chen2021}, LTE~\cite{lte-lee2022}), where a single model is trained and tested across different upsampling ratios.
\quad 
Our method is the first arbitrary-scale method to match (or surpass) fixed-scale performance across \textit{all} encoders in both single-pass inference and self-ensemble scenarios.

\input{tab/lightweight}
\subsection{Lightweight super-resolution}
\label{sec:lightweight-architectures}
The encoders and upsamplers considered so far are not designed to be run on devices with limited memory and power. Running super-resolution algorithms on mobile devices comes with unique challenges due to limited RAM and non-efficient support for certain common operations~\cite{Ignatov21Realtime}. The proposed CUF-based upsampler, in combination with a mobile-compatible encoder, can be effectively used for upscaling on mobile devices.

To test the effectiveness of the CUFs, we adopt a lightweight version of SwinIR ($0.9$M parameters) \cite{SWINIR_liang2021} as well as one of the best performing super-lightweight architectures for mobile devices ($30$K parameters) \cite{Du_2021_CVPR_abpn} as the encoder. For the latter, the original architecture (\Figure{abpn}, left) consists of a lightweight encoder and an upscaling module consisting of a series of convolutions, ReLU's, and a standard periodic shuffle layer. We replace this upscaling module with a CUF-based one (\Figure{abpn}, right). With instantiated CUFs, this means simply introducing a depthwise convolution followed by pointwise layers, where the weights are output by the learned CUFs.

\Table{lightweight} illustrates that CUFs provide the same level of accuracy while enabling continuous scale and efficient inference.
To our knowledge, this is the first mobile-friendly continuous scale super-resolution architecture.

\input{fig/abpn}

\subsection{Ablations}
\label{sec:ablations}
We run several ablation experiments in order to analyze the effects of various model choices. Here, we detail our choice of using DCT for positional encoding vs Fourier features, as well as an investigation of the redundancy of CUF filters versus subpixel convolution. An analysis on the various conditioning factors of our neural fields (positional encoding, kernel indices, scale, and target subpixel) can be found in the \SupplementaryMaterial.



\paragraph{Positional encoding}
\label{sec:posenc_results}
A large body of work uses a Fourier basis for the positional encodings, e.g. \cite{mildenhall2020nerf, vaswani2017attention}. However, we opt to use a DCT basis as this requires half of the sinusoidal projections and neurons at the hyper-network first layer. This trick has been used in other domains, such as random approximations of stationary kernel functions~\cite{rahimi2007random}.
\input{tab/sincos_dct}
In \autoref{tab:sincos} we show the PSNR, as obtained using a DFT versus a DCT basis, and verify that we do not observe loss in performance, while at the same time saving half of the computation associated with positional encoding operations.


\input{fig/low_rank}
\paragraph{Redundancy of filters}
As noted in Section~\ref{sec:analysis}, sub-pixel convolution does not enforce spatial correlation between neighboring weights in its convolutional filters, but has to learn them from data. CUF on the other hand generates the filters with a hypernetwork, which intrinsically builds on smoothness due to its functional form. Here, we test this hypothesis on the filters of CUF and sub-pixel convolutions. For this analysis, we use an EDSR encoder with $C_e=64$, $\kernelsize=3\times 3$, and $s=3$.
Subpixel convolution maps $C_e$ features to $s^2 C_e$ features, and each group of $s^2$ output features are rearranged to form the upscaled feature space. We therefore consider each group of $s^2$ filters of size $\kernelsize \times \kernelsize \times C_e$ and do an eigenvalue analysis to determine the compressibility of these filters.
This is done by forming $C_e$ separate matrices of size $s^2 \times C_e \kernelsize^2$ and performing PCA on each of them separately, forming $64$ separate sets of eigenvalues. We plot this distribution for trained and untrained sub-pixel layers in \Figure{subpixel_lowrank}. A faster decay means more redundancy, as more variance is explained by fewer components. In this case, training adds significant structure to the filters.

Conversely, each output channel in CUF is computed by $s^2$ filters of size $\kernelsize \times \kernelsize$. These are not directly comparable to subpixel convolution, but we perform an analogous eigenvalue analysis on trained and untrained CUFs.
As shown in \Figure{cuf_lowrank}, unlike subpixel convolution, CUFs actually become \textit{less} redundant with training, suggesting that the initial hypernetwork already imposes a very smooth prior on the filters.
In other words, CUF filters incorporate spatial correlations from the very beginning due to their functional form.


%% file: tab/params.tex
\begin{table}[t!]
\centering
\resizebox{\linewidth}{!}{
\begin{tabular}{llcc}
\toprule
Method & Composition & 
\multicolumn{2}{c}{Parameters (in M)}\\
& & encoder & spc. $4\times$
\\
\midrule
RDN & Convolutions & 22.00 & 0.30
\\
SwinT & Conv. and Self-Attention  & 11.60 & 0.30
\\ 
EDSR-baseline & Convolutions & ~~1.20 & 0.30
\\
SwinT-lightweight &Conv. and Self-Attention & ~~0.90 &  0.03
\\
ABPN &Convolutions & ~~0.03 & 0.03
\\ 
\bottomrule
\end{tabular}
} 
\caption{
\textbf{Encoders} --
Our experiments use CUF layers in combination with various encoders differing in composition and size.
Parameter counts reflect the encoder size and the upsampling head using a \subpix (spc) targeting $4\times$ upsampling.
} 
\label{tab:params}
\vspace{-1em}
\end{table}

%% file: fig/qualitative.tex
\begin{figure*}[t]
\centering
\footnotesize{
\begin{subfigure}{0.49\linewidth}
\setlength{\abovecaptionskip}{0pt}
    \begin{subfigure}[ht]{\linewidth}
        \begin{subfigure}[t]{0.32\linewidth}
            \includegraphics[width=\linewidth, height=0.625\linewidth]{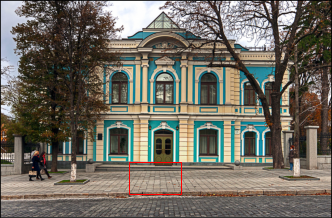}
            \caption*{Urban100 - \#008}
        \end{subfigure} 
        \begin{subfigure}[t]{0.32\linewidth}
            \includegraphics[width=\linewidth, height=0.625\linewidth]{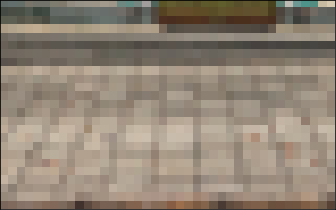}
            \caption*{LR}
        \end{subfigure}
        \begin{subfigure}[t]{0.32\linewidth}
            \includegraphics[width=\linewidth, height=0.625\linewidth]{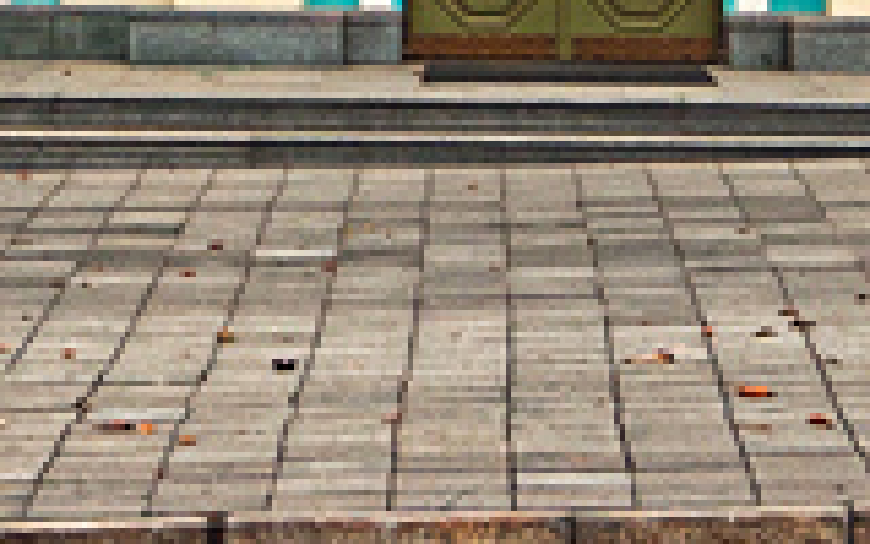}
            \caption*{GT}
        \end{subfigure}
    \end{subfigure}
    \begin{subfigure}[ht]{\linewidth}
        \begin{subfigure}{0.32\linewidth}
            \includegraphics[width=\linewidth, height=0.625\linewidth]{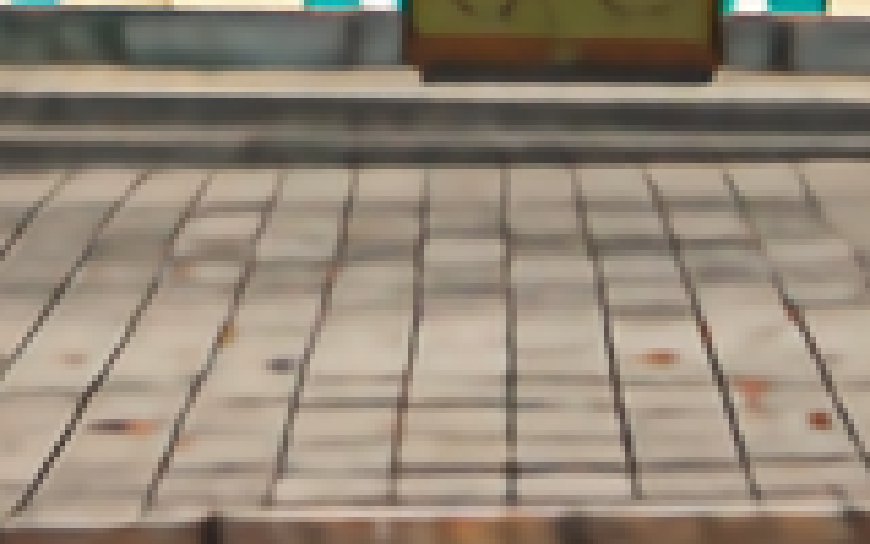} 
            \caption*{SwinIR + LIIF $\times$ 4}
        \end{subfigure}
        \begin{subfigure}{.32\linewidth}
            \includegraphics[width=\linewidth, height=0.625\linewidth]{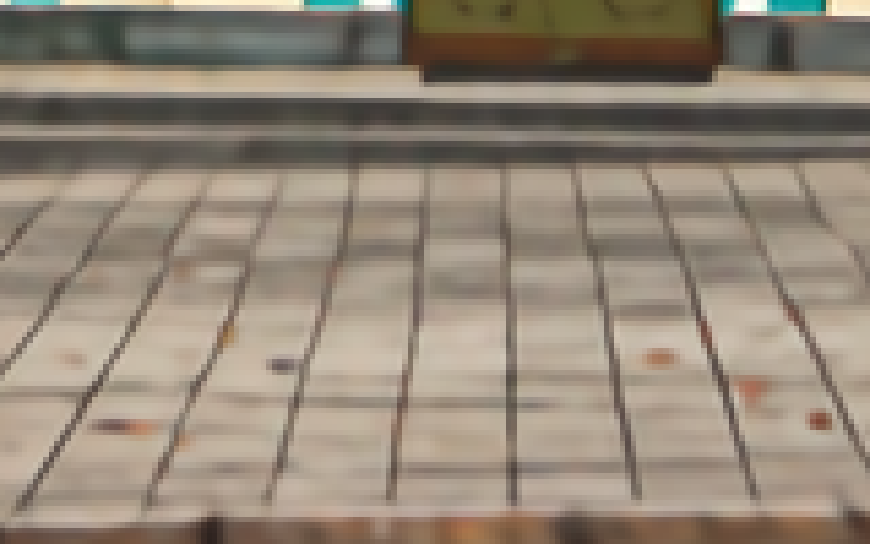}
            \caption*{SwinIR + LTE $\times$ 4}
        \end{subfigure}
        \begin{subfigure}{0.32\linewidth}
            \includegraphics[width=\linewidth, height=0.625\linewidth]{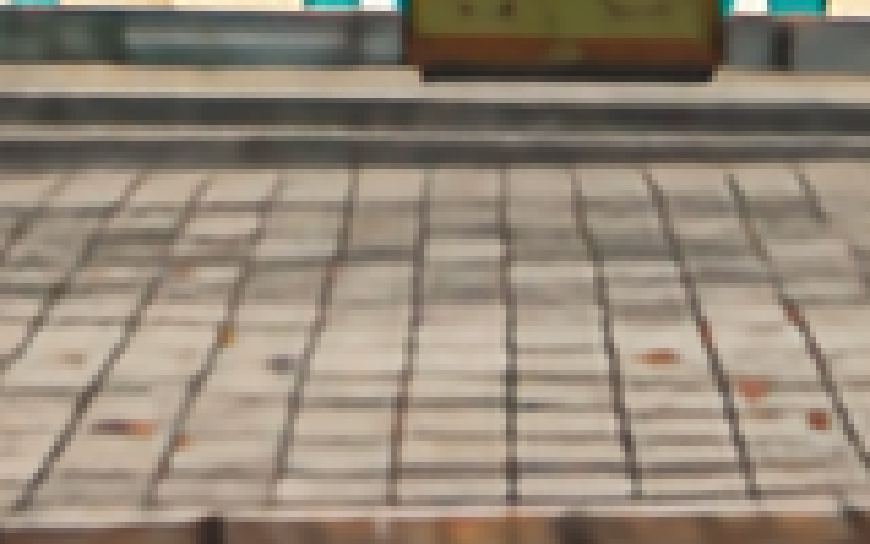} \caption*{SwinIR + CUF $\times$ 4}
        \end{subfigure}
    \end{subfigure}
\end{subfigure}
%
\begin{subfigure}{0.49\linewidth}
\setlength{\abovecaptionskip}{0pt} 
    \begin{subfigure}[ht]{\linewidth} 
        \begin{subfigure}[t]{0.32\linewidth}
            \includegraphics[width=\linewidth, height=0.625\linewidth]{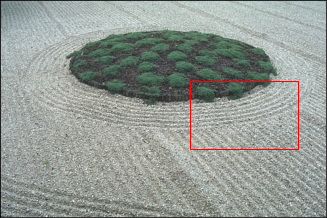}
            \caption*{B100 - \#86016}
        \end{subfigure} 
        \begin{subfigure}[t]{0.32\linewidth}
            \includegraphics[width=\linewidth, height=0.625\linewidth]{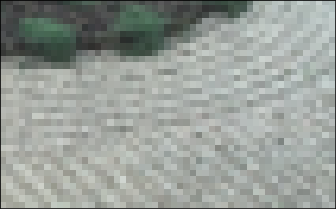}
            \caption*{LR}
        \end{subfigure}
        \begin{subfigure}[t]{0.32\linewidth}
            \includegraphics[width=\linewidth, height=0.625\linewidth]{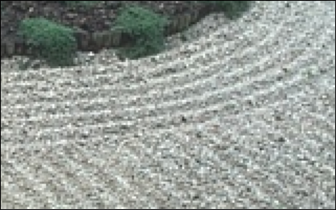}
            \caption*{GT}
        \end{subfigure}
    \end{subfigure}
    \begin{subfigure}[ht]{\linewidth}
        \begin{subfigure}{0.32\linewidth}
            \includegraphics[width=\linewidth, height=0.625\linewidth]{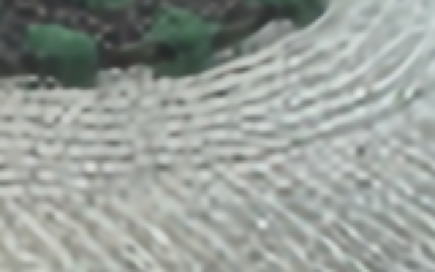} 
            \caption*{EDSR + LIIF $\times$ 3}
        \end{subfigure}
        \begin{subfigure}{.32\linewidth}
            \includegraphics[width=\linewidth, height=0.625\linewidth]{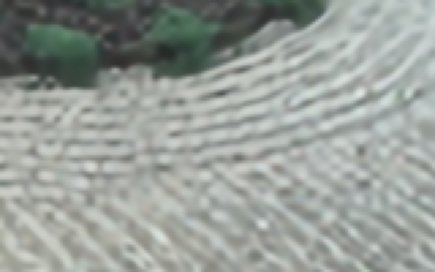}
            \caption*{EDSR + LTE $\times$ 3}
        \end{subfigure}
        \begin{subfigure}{0.32\linewidth}
            \includegraphics[width=\linewidth, height=0.625\linewidth]{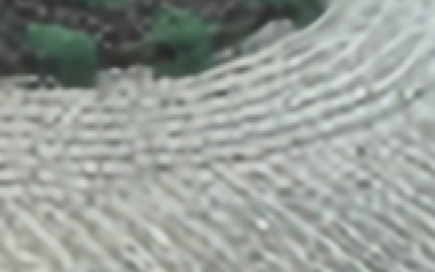} \caption*{EDSR + CUF $\times$ 3}
        \end{subfigure}
    \end{subfigure}
\end{subfigure}
\begin{subfigure}{0.49\linewidth}
\setlength{\abovecaptionskip}{0pt} 
    \begin{subfigure}[ht]{\linewidth} 
        \begin{subfigure}[t]{0.32\linewidth}
            \includegraphics[width=\linewidth, height=0.625\linewidth]{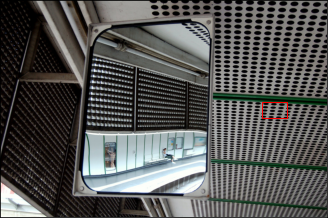}
            \caption*{Urban100 - \#030}
        \end{subfigure} 
        \begin{subfigure}[t]{0.32\linewidth}
            \includegraphics[width=\linewidth, height=0.625\linewidth]{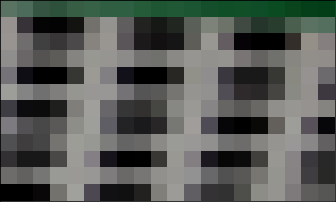}
            \caption*{LR}
        \end{subfigure}
        \begin{subfigure}[t]{0.32\linewidth}
            \includegraphics[width=\linewidth, height=0.625\linewidth]{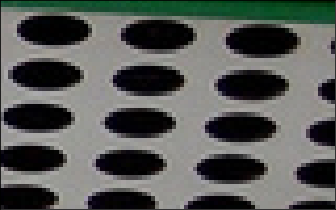}
            \caption*{GT}
        \end{subfigure}
    \end{subfigure}
    \begin{subfigure}[ht]{\linewidth}
        \begin{subfigure}{0.32\linewidth}
            \includegraphics[width=\linewidth, height=0.625\linewidth]{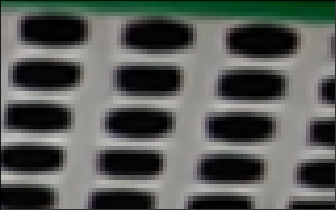} 
            \caption*{EDSR + LIIF $\times$ 4}
        \end{subfigure}
        \begin{subfigure}{.32\linewidth}
            \includegraphics[width=\linewidth, height=0.625\linewidth]{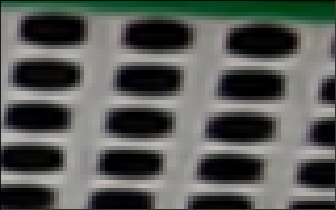}
            \caption*{EDSR + LTE $\times$ 4}
        \end{subfigure}
        \begin{subfigure}{0.32\linewidth}
            \includegraphics[width=\linewidth, height=0.625\linewidth]{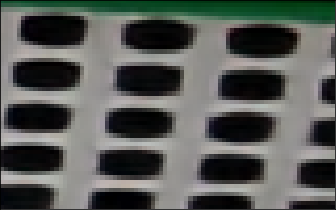} \caption*{EDSR + CUF $\times$ 4}
        \end{subfigure}
    \end{subfigure}
\end{subfigure} 
\begin{subfigure}{0.49\textwidth}
\setlength{\abovecaptionskip}{0pt} 
    \begin{subfigure}[ht]{\linewidth}
        \begin{subfigure}[t]{0.32\linewidth}
            \includegraphics[width=\linewidth, height=0.625\linewidth]{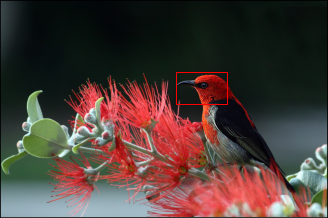}
            \caption*{Urban100 - \#008}
        \end{subfigure} 
        \begin{subfigure}[t]{0.32\linewidth}
            \includegraphics[width=\linewidth, height=0.625\linewidth]{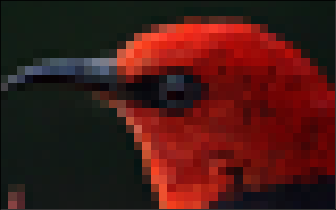}
            \caption*{LR}
        \end{subfigure}
        \begin{subfigure}[t]{0.32\linewidth}
            \includegraphics[width=\linewidth, height=0.625\linewidth]{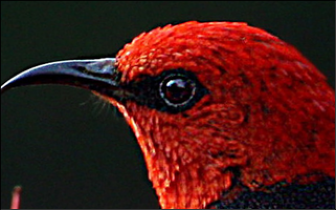}
            \caption*{GT}
        \end{subfigure}
    \end{subfigure}
    \begin{subfigure}[ht]{\linewidth}
        \begin{subfigure}{0.32\linewidth}
            \includegraphics[width=\linewidth, height=0.625\linewidth]{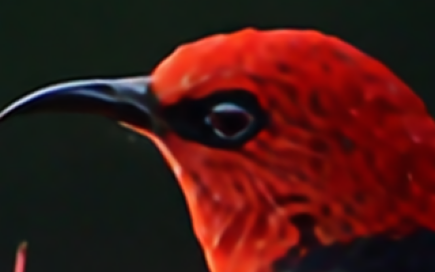} 
            \caption*{RDN + LIIF $\times$ 8}
        \end{subfigure}
        \begin{subfigure}{.32\linewidth}
            \includegraphics[width=\linewidth, height=0.625\linewidth]{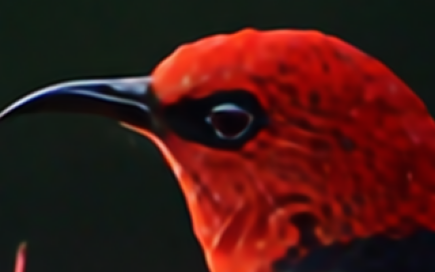}
            \caption*{RDN + LTE $\times$ 8}
        \end{subfigure}
        \begin{subfigure}{0.32\linewidth}
            \includegraphics[width=\linewidth, height=0.625\linewidth]{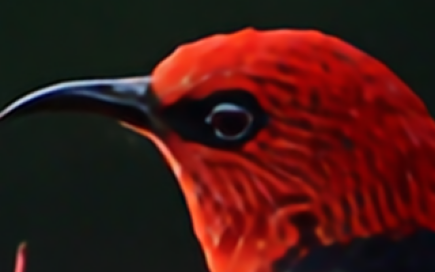} \caption*{RDN + CUF $\times$ 8}
        \end{subfigure}
    \end{subfigure}
\end{subfigure} 
} 
\vspace{-.5em}
\caption{
\textbf{Qualitative evaluation} --  CUF combined with different encoders and upsampling scales.
\label{fig:qualitative}}
\vspace{-1em}
\end{figure*}

%% file: tab/lightweight.tex
\begin{table*}
\centering
\resizebox{\textwidth}{!}{
\footnotesize
\begin{tabular}{ll||c|c|c|c|c|c|c|c|c|c|c|c|c|c|c|c}
 \multicolumn{18}{c}{Multi scale up-sampling methods} \\
 \hline
  &  &
  \multicolumn{4}{c|}{Set5} &
  \multicolumn{4}{c|}{Set14} &
  \multicolumn{4}{c|}{BSD100} & 
  \multicolumn{4}{c}{Urban100}  
  \\
 Encoder & Upsampler & \tiny{$\times$ 2} &   \tiny{$\times$ 3} &   \tiny{$\times$ 4} &  \tiny{$\times$ 6} & 
  \tiny{$\times$ 2} &   \tiny{$\times$ 3} &   \tiny{$\times$ 4} & \tiny{$\times$ 6} &   
    \tiny{$\times$ 2} &   \tiny{$\times$ 3} &   \tiny{$\times$ 4} & \tiny{$\times$ 6} &   
      \tiny{$\times$ 2} &   \tiny{$\times$ 3} &   \tiny{$\times$ 4} & \tiny{$\times$ 6}  
\\\midrule
SwintIR-light \cite{SWINIR_liang2021}& \subp &
38.14 
& 34.62 
& 32.44 
&-- 
& 33.86 
& 30.54 
& 28.77 
&-- 
& 32.31 
& 29.20 
& 27.69 
&-- 
& 32.76 
& 28.66 
& 26.47 
&-- 
\\
& \textbf{CUF} (ours)   
& 38.17
& 34.62
& 32.46
& 29.09
& 33.93
& 30.57
& 28.83
& 26.61
& 32.30
& 29.23
& 27.71
& 25.95
& 32.73
& 28.68
& 26.55
& 24.11 
\\ 
\hline 
ABPN\cite{Du_2021_CVPR_abpn} & \subp 
&37.30&
33.39&
31.11&--&
32.90&
29.68&
27.97&--&
31.67&
28.64&
27.14&--&
30.37&
26.77&
24.96&--
\\
& \textbf{CUF} (ours)
&
37.35 &
33.55 &
31.31 &
28.06 &
32.96 &
29.74 &
28.04 &
25.86 &
31.72 &
28.69 &
27.19 &
25.52 &
30.53 &
26.92 &
25.05 &
23.01 
\end{tabular}
} 
\caption{
\textbf{Lightweight super-res} --
Our CUFs enable continuous super-resolution with lightweight (SwinIR-lightweight \cite{SWINIR_liang2021}) and mobile-friendly (ABPN \cite{Du_2021_CVPR_abpn}) backbones without quality loss.
}
\label{tab:lightweight}
\vspace{-1em}
\end{table*} 

%% file: fig/abpn.tex
\begin{figure}[b]
\begin{center}
\includegraphics[width=.99\columnwidth]{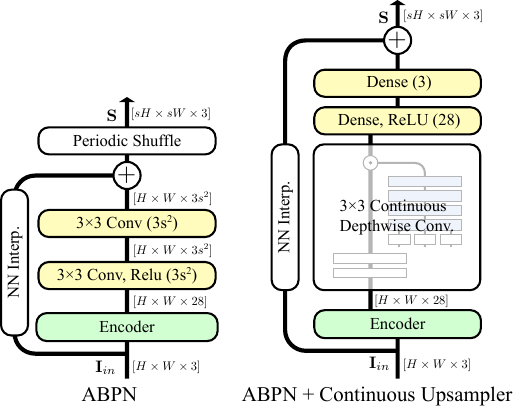}
\end{center}
\caption{
\textbf{Architectures (lightweight)} --
The original ABPN~\cite{Du2Anchorbased} single scale architecture (left) and our arbitrary-scale version using the implicit upsampler (right). Note how the two consecutive $3 \times 3$ native convolutions are exchanged for a $3 \times 3$ depth-wise implicit convolution followed by point-wise layers.}
\label{fig:abpn}
\vspace{-2em}
\end{figure}

%% file: tab/sincos_dct.tex
\begin{table}
\centering
\resizebox{.5\linewidth}{!}{ 
\begin{tabular}{@{}l|ccc@{}}
&$2\times$ & $3\times$ & $4\times$  
\\
\midrule
Fourier &
30.45 & 
26.85 & 
25.00 
\\
DCT  & 
30.44 & 
26.86 &  
25.00 
\end{tabular}
} 
\caption{
\textbf{Cosine basis} -- matching results but using half of sinusoidal projections and neurons at the hyper-network input.
PSNR from Urban-100 dataset using a ABPN  encoder.}
\label{tab:sincos}
\vspace{-1em}
\end{table}

%% file: fig/low_rank.tex
\begin{figure}[b]
\begin{center}
    \begin{subfigure}[ht]{0.49\columnwidth}
        \includegraphics[width=\textwidth]{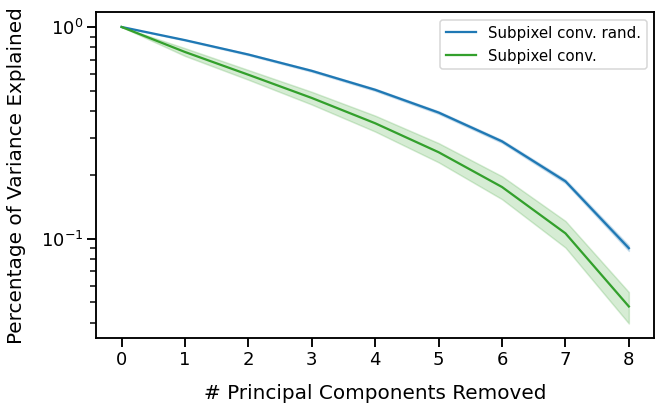}
        \caption{Subpixel Convolutions}
        \label{fig:subpixel_lowrank}
    \end{subfigure}
    \begin{subfigure}[ht]{0.49\columnwidth}
        \includegraphics[width=\textwidth]{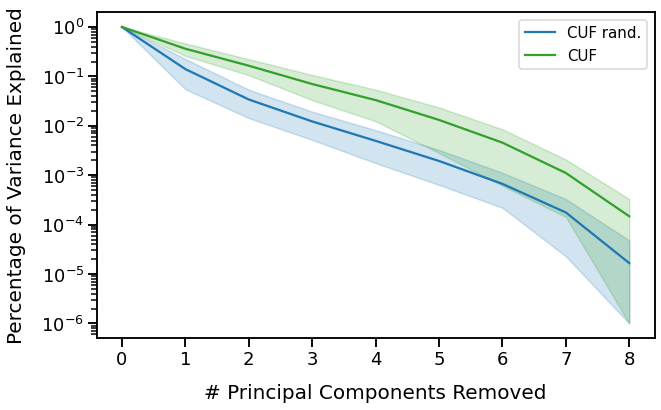}
        \caption{CUF}
        \label{fig:cuf_lowrank}
    \end{subfigure}
\caption{
CUF depthwise continuous filters are \textbf{low rank by construction}. Subpixel-convolution fails to find low rank filters.}
\label{fig:lowrank}
\end{center}
\vspace{-1em}
\end{figure}

%% file: sec/5_conclusions.tex
\vspace{-2pt}
\section{Conclusions} 
\vspace{-2pt}
We propose CUF, a computationally efficient modeling of continuous upsampling filters as neural fields. 
Different from previous arbitrary-scale architectures, the performance gains obtained by our upsampler are not due to increased capacity but rather to a more efficient use of model parameters. 
A single hyper-network supports filter adaptation across scales while at the same time CUF's upsampling head has fewer parameters than previous arbitrary-scale heads, as well as fewer parameters than a single \subpix head at $4 \times$ upsampling.
CUF is the first arbitrary scale model that 
can be effectively used for upscaling on mobile devices, an area previously dominated by single scale \subpix architectures. 
\quad
CUF's archtecture is quite general. A natural direction of future work is to investigate its use for architectures with multiple upsampling layers (and the reuse of filters across scales)
and for different image generation settings such as GANs~\cite{goodfellow2020generative} and diffusion models~\cite{ho2020denoising}. 
\quad
Neural fields have done particularly well on higher-dimensional signals. Another line of future work is the use of CUF for efficient super-resolution of video data.

%% file: sec/X_supplementary.tex
\appendix


\twocolumn[
\centering
\Large
\textbf{CUF: Continuous Upsampling Filters} \\
\vspace{0.5em}Supplementary Material \\
\vspace{1.0em}
] 
\appendix

\section{Experiment details}
\paragraph{Training Hyper-parameters} 
\input{tab/hyper_params}
\autoref{tab:hyper} contains the hyper-parameters used for training the ablated models, taken from their single-scale (\subp) training settings. All models are trained for $1K$ epochs with  $\mathcal{L}_1$ loss and ADAM optimizer by setting $\beta_1=0.9$,
$\beta_2=0.999$,
and $\epsilon=1e^{-8}$. We use a step-wise learning rate schedule that is halved at epochs $[500,800,900,950]$. Unless referred as single-scale, models were trained with random scales by sampling the scale factor uniformly within the continuous interval $[1, 4]$.
In order to ensure that the dimensions within the training mini-batch match despite heterogeneous scale factors, 
we first scale each image as required and then apply the same crop size to both LR and HR images, such that the HR random crop contains $1/s^2$ of the content of the LR crop, and fix the relative grid coordinates to point to the random subregion.

\input{tab/cuf_hyper_params}
\paragraph{CUF's hyper parameters}
\label{sup:cuf_hyper_params}
\autoref{tab:hyper_CUF} describes the number of neurons  used in the presented ablations across different encoders.
$C_e$ was chosen to replicate the original number of output features of each encoder.
CUF's hyper-parameters were obtained by grid search on the EDSR-baseline and Set5 dataset and replicated on the remaining encoders.
CUF's positional encoding hyper-parameters enforce a small number of basis per input parameter. For each input (represented in $2D$ space), the number of basis $N^2$ was searched within the set $\{0, 1^2, 2^2, ..., 5^2\}$ while $f_\text{max}$ within the set $\{0, 0.5, ...,4.0\}$ for $\posenc(\relative^\scale(\x))$ and $\posenc(\scale)$, and within the set $\{0, 0.5,.., 3.0\}$ for 
$\posenc(\offset)$.
The final positional encoding hyper-parameters adopted are 
$\posenc(\relative^\scale(\x);):\{N^2=25; f_\text{max}=2.0\}$,
$\posenc(\scale):\{N^2=25; f_\text{max}=2.0\}$
and
$\posenc(\offset):\{N^2=9; f_\text{max}=1.0\}$.

\section{On the use of ensemble}
\label{sup:ensemble}

\input{tab/on_ensemble}

The baseline settings from  LIIF \cite{LIIF_chen2021} and LTE \cite{lte-lee2022}  include locally ensembling pixels around the target sub-pixel, by shifting by its position by half pixel in the low resolution grid, and averaging their results. This procedure introduces a training and inference overhead, as the sampled points is increased by a factor of four. 
As a direct consequence,
 during training models adopting local self-ensemble evaluate four times more gradients per optimization step. 
In order to disentangle possible optimization side effects,  
in this section we ablate the models LIIF and LTE under same optimization conditions as other models, that is, no ensemble is adopted during training, but on inference time only.
LTE presents a strong result on scales $3$ and larger, but a reduction in performance on scale $2$, in which LIIF matches or surpass its performance. 
Overall, this ablation confirms the benefits from CUF as the lighter arbitrary scale upsampler with strong performance under single pass and self-ensemble settings, across both smaller and larger scales.

\section{On the use of positional encoding}
\label{sup:posenc}
\input{fig/posenc_versus_raw}

\autoref{fig:posenc_raw} contains a comparison of CUF models trained using the neural-fields parameters as raw values versus the projection using positional encoding.
The stronger impact of using positional encoding is observed on ABPN encoder (\autoref{fig:qualitative_posenc}) and Urban-100 dataset \cite{Urban100_Huang2015}. We note that the content of this dataset is characterized by sharp straight lines and geometric structures, thus
the quantitative gain is aligned with the expected behaviour.

\section{Conditioning on the kernel indexes}
\label{sup:kin_kout}
\input{fig/k_in_out}
\autoref{fig:k_in_out} contains a comparison between representing the kernel indexes $k_i, k_j$ 
as input parameters to the neural-fields 
 versus representing their discrete set as individual neurons at the output of the hyper-network.
On the efficiency side, setting them as the hyper-network output neurons reduces memory and computation used, as hidden layers are shared.
On the other hand, the layers of the hyper-network and its nonlinearities provide additional expressiveness compared to the linear transformation 
 used in the multi-headed version. This additional expressiveness results in in performance improvement in stronger encoders (RDN, SWINIR), but not on smaller ones (ABPN, EDSR). Thus, we recommend conditioning on kernel indexes only for those encoders that take advantage of it. 

\section{\subpix vs. \cufi}
\label{sup:ops}

In this section we compare the costs associated with \subpix and \cufi upsampling heads.
The presented comparison contrasts their designs choices based on full convolution (\subpix) versus depthwise-pointwise decomposition (CUF).
As unitary element of comparison we evaluate the number of multiplications performed to produce a single output pixel.
We assume an input feature map with $C_{in}$ channels, the resulting image with $C_{out}$ channels and that both \subpix and \cufi adopt filters of same size $\kernelsize$.

 \cufi architecture is composed with a depthwise convolution and pointwise projections. Its three layers perform respectively $C_{in}*k^2$, $C_{in}*C_{in}$ and $C_{in}*C_{out}$ multiplications per output pixel.

Next, we cover two common compositions with \subpix. The most common design for upsampler heads targeting high quality results is to combine a \subpix layer with a pointwise layer projecting from $C_{in}$ into RGB channels ($C_{out}$)
(\cite{EDSR_Lim_2017,RDN_zhang2020,SWINIR_liang2021}). 
In this setting, both \subp and \cufi have identical output layers, that is removed from our analysis.
The number of multiplications performed by the \subpix layer alone per target subpixel is: $C_{in}*k*k*C_{in}$.
Thus, the fraction of multiplications performed by \cufi in relation to \subpix can be expressed as:
$\frac{k^2 +C_{in}}{k^2*C_{in}} = \frac{1}{C_{in}} + \frac{1}{k^2}$. 
That is, the decomposition has the desired effect of saving computation whenever $C_{in},k>1$.

An alternative use of a \subpix layer is its direct use as output layer. 
In this case, the three layers that compose CUF's upsampling head are compared to the full expansion convolution alone. 
 Thus, the total operations performed by \cufi is smaller that those performed by \subpix whenever $k^2 +C_{in} +C_{out} < k^2*C_{out} $. 

In practice, the depthwise-pointwise decomposition adopted by \cufi faces a memory drawback of storing an extra feature map created in-between the decomposition layers $C_{in}$ (\Figure{teaser}). The reduction of such drawback with fused-convolutions is left as future work \cite{fuse}.

\section{Qualitative comparisons}
\label{sup:qualitative}

The difference between existent arbitrary-scale up-samplers can only be observed at textured regions of the image. In this section we disentangle the rule of the encoder and upsampler in the perceived quality of the results. 
\Figure{qual_fractional} contain additional results, with non-integer scale factors.

\input{fig/qualitative_8x}
\input{fig/fractional_scale}

%% file: tab/hyper_params.tex
\begin{table}[b]
\centering
\resizebox{0.7\linewidth}{!}{ 
\begin{tabular}{@{}lccc@{}}
\toprule
Encoder & Batch & Crop & Initial LR\\
\midrule
EDSR-baseline \cite{EDSR_Lim_2017}& 16 & 48 & $1e^{-4}$ \\
RDN \cite{RDN_zhang2020}& 16  & 48 & $1e^{-4}$ \\
SWINIR \cite{SWINIR_liang2021}& 32  & 48 & $2e^{-4}$ \\
SWINIR-lightweight \cite{SWINIR_liang2021}& 64  & 64& $2e^{-4}$ \\
ABPN \cite{Du_2021_CVPR_abpn}& 16 & 64 & $1e^{-3}$  \\
\bottomrule
\end{tabular}
} 
\caption{
\textbf{Training hyper-parameters}: replicate each encoder's original setting.}
\label{tab:hyper}
\end{table}

%% file: tab/cuf_hyper_params.tex
\begin{table}[b]
\centering
\resizebox{0.7\linewidth}{!}{ 
\begin{tabular}{@{}lccccccccc@{}}
\toprule
Encoder & 
$C_e$ & 
$C_h$ & Params(K)
%
\\
\midrule
EDSR-baseline  &
64 & 32 & 10\\
RDN 
\\
SWINIR \\
\hline
SWINIR-lightweight &
60 & 32 & 10
\\ 
\hline
ABPN  (k as input)&
28 & 28  & 5
\\
ABPN  (k as output)& & & 11
\\
\end{tabular}
} 
\caption{
\textbf{CUF's hyper-parameters}: $C_e$ chosen as each the encoder output features.}
\label{tab:hyper_CUF}
\end{table}

%% file: tab/on_ensemble.tex
\begin{table}[ht!]
\centering
\resizebox{\linewidth}{!}{
\begin{tabular}{llc||c|c|c|c|c} 
\hline
 \multicolumn{7}{c}{Multi-scale up-sampling methods - DIV2k} \\
 \hline
 Encoder & Upsampler & Ens.& \multicolumn{3}{c|}{seen scales}& \multicolumn{2}{c}{unseen scales} \\
 & & & \tiny{$\times$ 2} &   \tiny{$\times$ 3} &   \tiny{$\times$ 4} &  \tiny{$\times$ 6} &  
 \tiny{$\times$ 12} 
\\
\hline 
EDSR-baseline \cite{EDSR_Lim_2017}& Sub-pixel conv.
& & \second 34.69 & \second 30.94 & \second 28.97 & --&--
\\
 & LIIF & & 34.63&30.95&28.97&26.72
 & 23.66 
 \\
 & LTE & & 
 34.63 & \first 30.99 & \first 29.01 & \first26.77 &
 \first 23.74\\
 
 & CUF (ours) 
&
&\first 34.70
&\first 30.99
&\first 29.01
&\second 26.76
&\second 23.73 
 \\ 

\hline
& Sub-pixel conv. & +$_\text{geo}$&
\second 34.78 & 31.03&29.06&--&--
\\

& LIIF& +$_\text{geo}$&
34.74 & \second31.05  & 29.07 & 
26.80 & 23.76 
\\
 & LTE & +$_\text{geo}$&
 34.72 & \first31.07 &
 \second29.08 & \first26.83 & 
 \first23.79 
 \\
& CUF (ours) & +$_\text{geo}$
&\first34.79 &\first31.07
&\first29.09 &\second26.82 &\second23.78 
\\

 \end{tabular}}
\caption{
\textbf{Disentangling the effect of ensembling on optimization}: Models trained under same supervision (no ensemble), and tested with (marked with $+_{geo}$) and without geometric self-ensemble. Results on DIV2K's validation subset~\cite{DIV2k_Timofte2017}.
}\label{tab:ensembling}
\end{table}

%% file: fig/posenc_versus_raw.tex
\begin{figure}
\begin{center}
\includegraphics[width=.99\columnwidth]{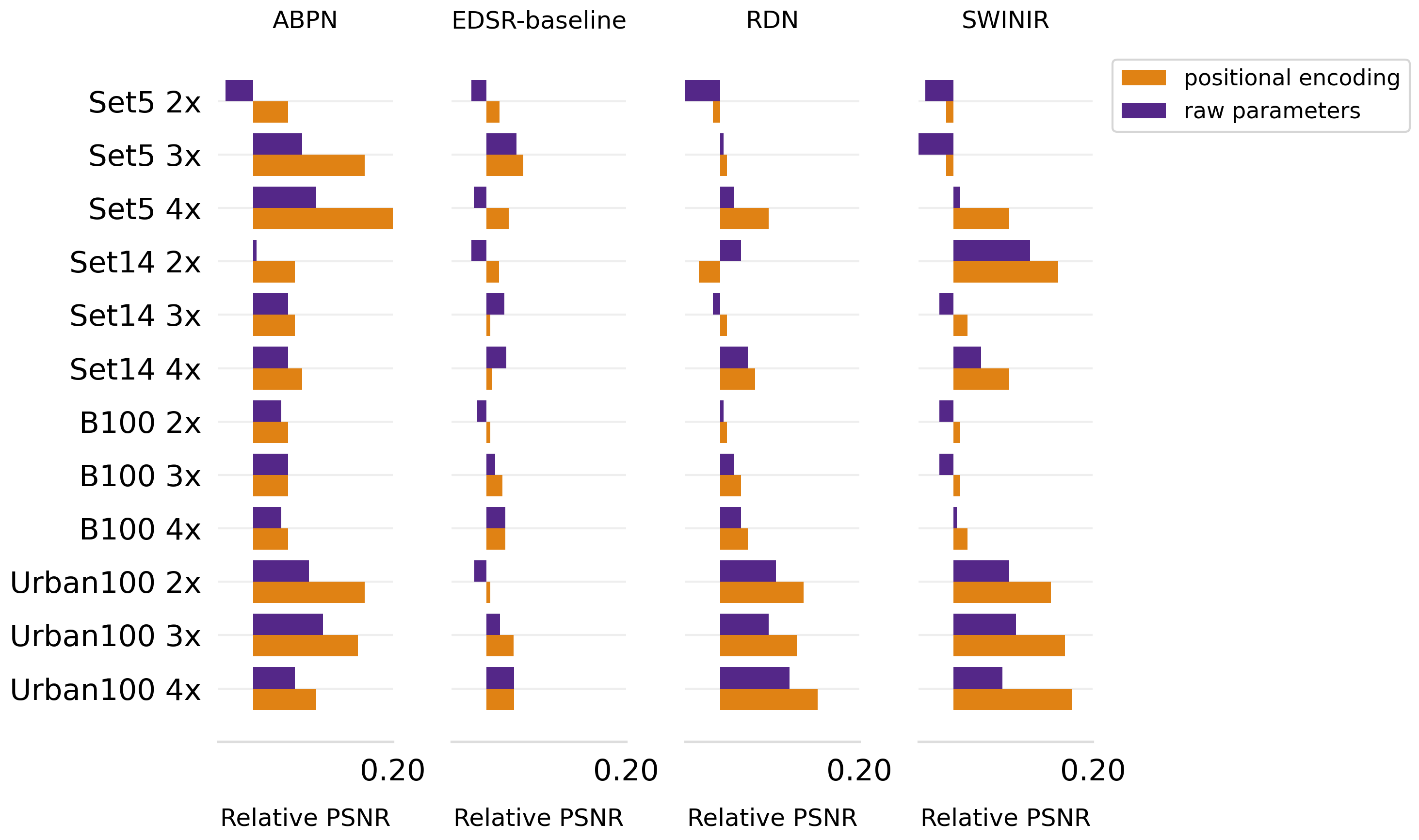}
\end{center}
\caption{Impact of CUF's
\textbf{positional encoding} on different datasets and encoders.
Bar plots represent PSNR differences relative to baseline models adopting sub-pixel convolutions and corresponding encoder.}
\label{fig:posenc_raw}
\end{figure}

\begin{figure}[t]
\begin{center}
\begin{subfigure}{\linewidth}
\setlength{\abovecaptionskip}{0pt}
    \begin{subfigure}[ht]{\linewidth}
        \begin{subfigure}[t]{0.32\linewidth}
            \includegraphics[width=\linewidth]{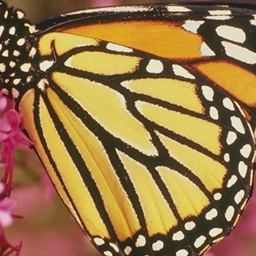}
            \caption*{HR}
        \end{subfigure} 
        \begin{subfigure}[t]{0.32\linewidth}
            \includegraphics[width=\linewidth]{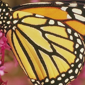}
            \caption*{LR}
        \end{subfigure}
        \hfill
    \end{subfigure}
    \begin{subfigure}[ht]{\linewidth}
        \begin{subfigure}{0.22\linewidth}
            \includegraphics[width=\linewidth]{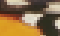} 
            \caption*{LR crop}
        \end{subfigure}
        \begin{subfigure}{.22\linewidth}
            \includegraphics[width=\linewidth]{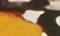}
            \caption*{w/o. pos. enc.}
        \end{subfigure}
        \begin{subfigure}{0.22\linewidth}
        \includegraphics[width=\linewidth]{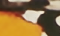} \caption*{with. pos. enc.}
        \end{subfigure}
        \begin{subfigure}{0.22\linewidth}
            \includegraphics[width=\linewidth]{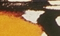} 
            \caption*{GT ($\times$ 3)}
        \end{subfigure}
    \end{subfigure}

    \begin{subfigure}[ht]{\linewidth}
        \begin{subfigure}{0.22\linewidth}
            \includegraphics[width=\linewidth]{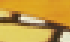} 
            \caption*{LR crop}
        \end{subfigure}
        \begin{subfigure}{.22\linewidth}
            \includegraphics[width=\linewidth]{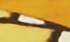}
            \caption*{w/o. pos enc}
        \end{subfigure}
        \begin{subfigure}{0.22\linewidth}
        \includegraphics[width=\linewidth]{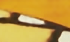} \caption*{with. pos enc}
        \end{subfigure}
        \begin{subfigure}{0.22\linewidth}
            \includegraphics[width=\linewidth]{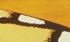} 
            \caption*{GT ($\times$ 3)}
        \end{subfigure}
    \end{subfigure}
\end{subfigure}
%
\end{center}
\caption{
\textbf{Qualitative evaluation on mobile-compatible encoder} --  ABPN-CUF with and without positional encoding. Butterfly image from Set5 dataset. 
}
\label{fig:qualitative_posenc}
\end{figure}

%% file: fig/k_in_out.tex
\begin{figure}
\begin{center}
\includegraphics[width=.99\columnwidth]{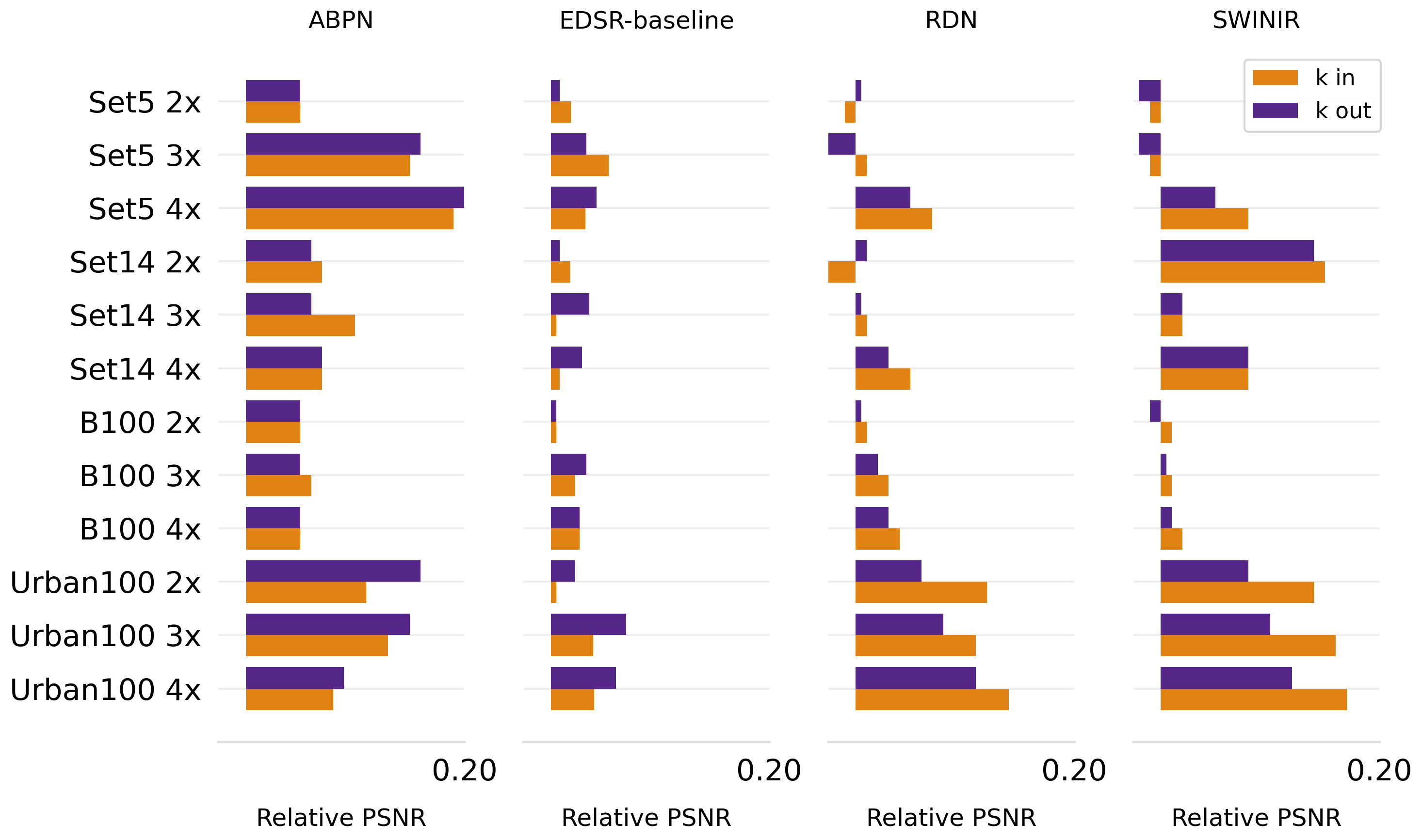}
\end{center}
\caption{
\textbf{Comparison between conditioning the neural-fields on the kernel indexes $(k_i,k_j)$ versus its discretization at the hyper-network output layer.} Stronger encoders take advantage of the hyper-network depth and non linearities. 
Bar plots represent PSNR differences relative to baseline models adopting sub-pixel convolutions and corresponding encoder.}
\label{fig:k_in_out}
\end{figure}

%% file: fig/qualitative_8x.tex
\begin{figure*}[ht!]
\centering
\begin{subfigure}{.48\linewidth}
\vfill
    \begin{subfigure}[ht]{0.99\textwidth}
        \vfill
        \begin{subfigure}[t]{0.32\textwidth}
        \includegraphics[width=\linewidth, height=0.625\linewidth]{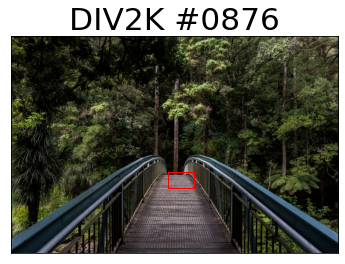} 
        \end{subfigure} 
        \hfill
        \begin{subfigure}[t]{0.32\textwidth}
        \includegraphics[width=\linewidth, height=0.625\linewidth]{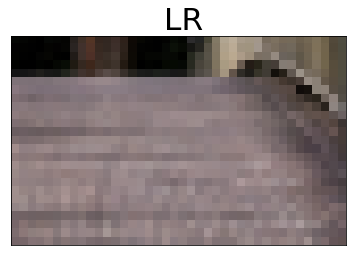}
        \end{subfigure}
        \hfill
        \begin{subfigure}[t]{0.32\textwidth}
        \includegraphics[width=\linewidth, height=0.625\linewidth]{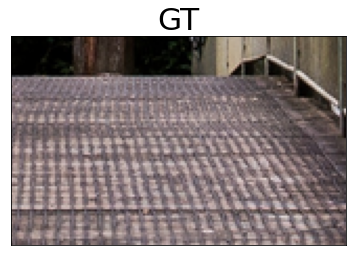}
        \end{subfigure}
    \end{subfigure}
    \begin{subfigure}[ht]{0.99\textwidth}
        \includegraphics[width=\linewidth, height=0.625\linewidth]{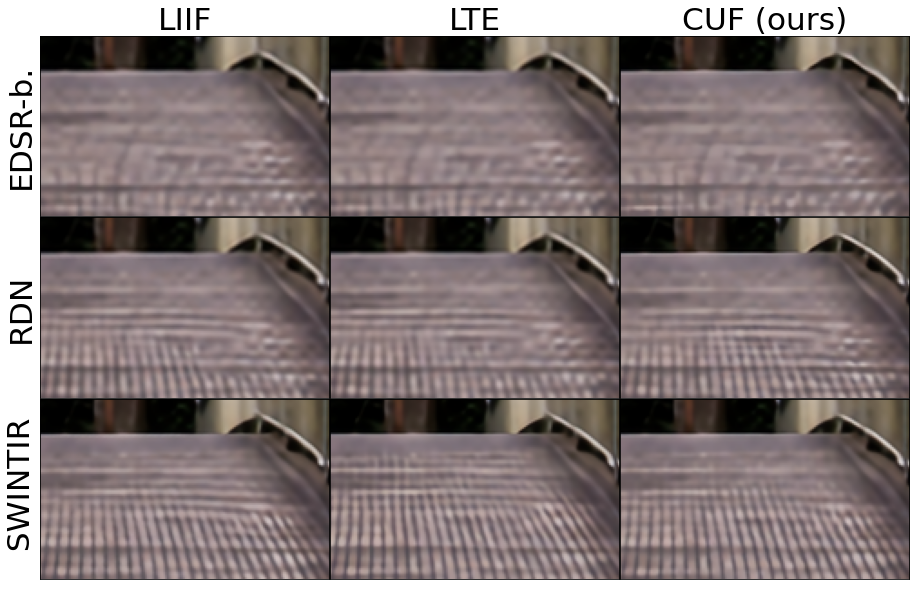}
    \end{subfigure}
\end{subfigure}
\hfill
\begin{subfigure}{.48\linewidth}
\vfill
    \begin{subfigure}[ht]{0.99\textwidth}
        \vfill
        \begin{subfigure}[t]{0.32\textwidth}
        \includegraphics[width=\linewidth, height=0.625\linewidth]{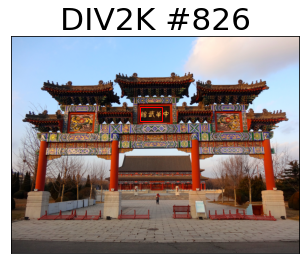} 
        \end{subfigure} 
        \hfill
        \begin{subfigure}[t]{0.32\textwidth}
        \includegraphics[width=\linewidth, height=0.625\linewidth]{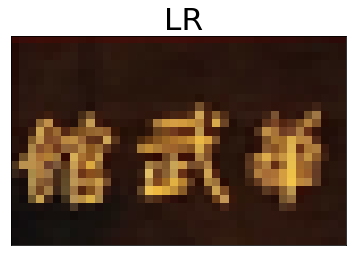}
        \end{subfigure}
        \hfill
        \begin{subfigure}[t]{0.32\textwidth}
        \includegraphics[width=\linewidth, height=0.625\linewidth]{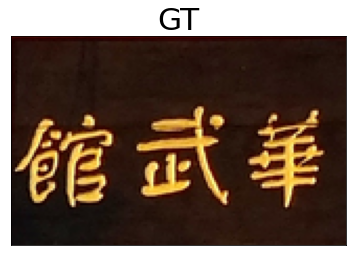}
        \end{subfigure}
    \end{subfigure}
    \begin{subfigure}[ht]{0.99\textwidth}
        \includegraphics[width=\linewidth, height=0.625\linewidth]{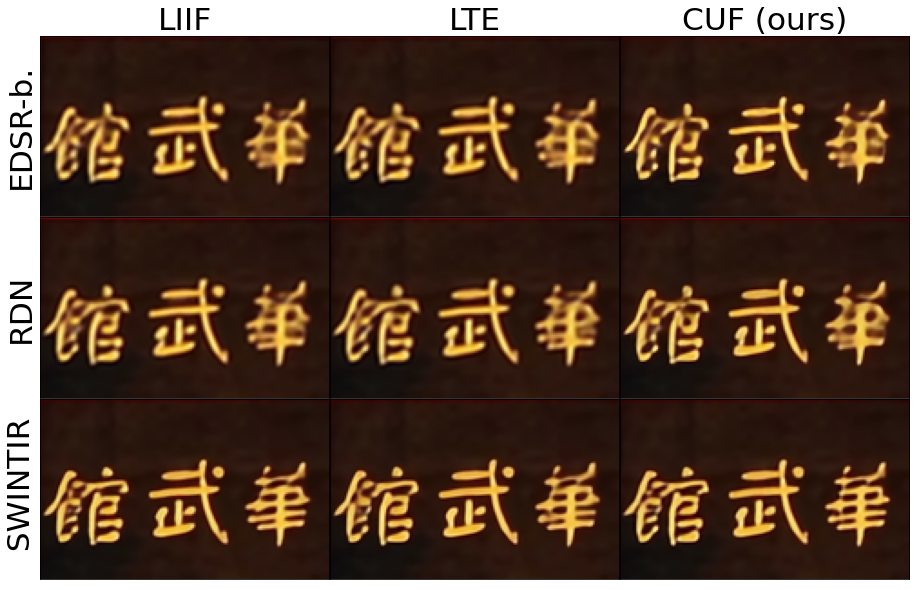}
    \end{subfigure}
\end{subfigure}

\begin{subfigure}{.48\linewidth}
\vfill
    \begin{subfigure}[ht]{0.99\textwidth}
        \vfill
        \begin{subfigure}[t]{0.32\textwidth}
        \includegraphics[width=\linewidth, height=0.625\linewidth]{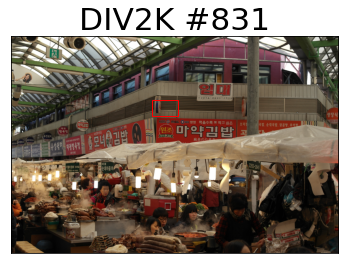} 
        \end{subfigure} 
        \hfill
        \begin{subfigure}[t]{0.32\textwidth}
        \includegraphics[width=\linewidth, height=0.625\linewidth]{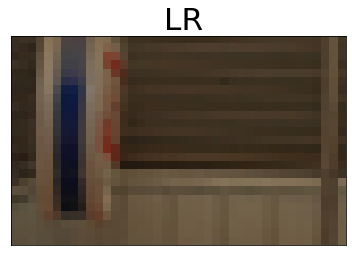}
        \end{subfigure}
        \hfill
        \begin{subfigure}[t]{0.32\textwidth}
        \includegraphics[width=\linewidth, height=0.625\linewidth]{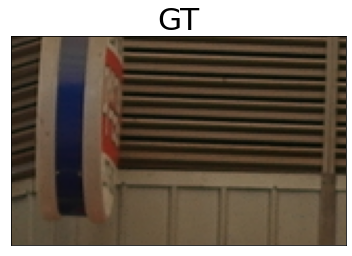}
        \end{subfigure}
    \end{subfigure}
    \begin{subfigure}[ht]{0.99\textwidth}
        \includegraphics[width=\linewidth, height=0.625\linewidth]{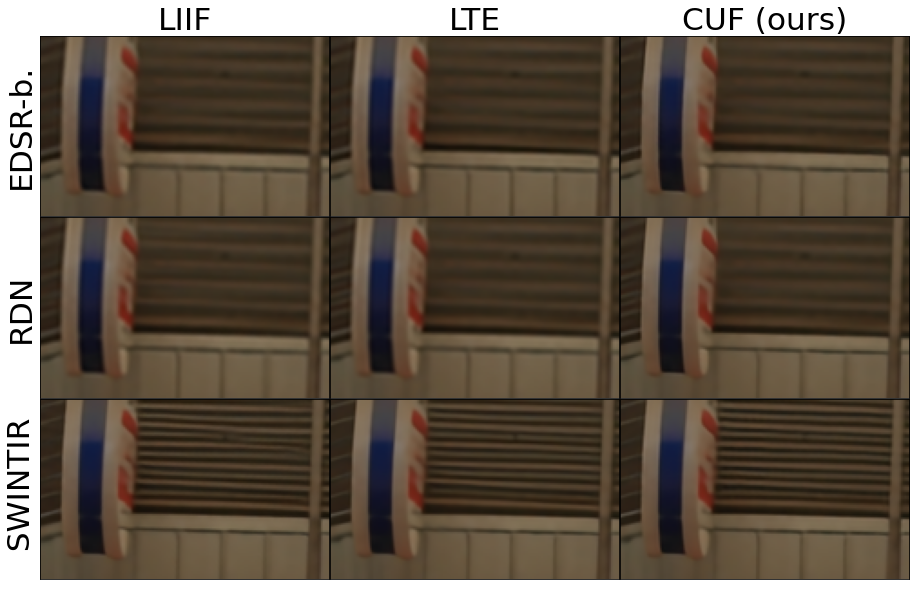}
    \end{subfigure}
\end{subfigure}
\hfill
\begin{subfigure}{.48\linewidth}
\vfill
    \begin{subfigure}[ht]{0.99\textwidth}
        \vfill
        \begin{subfigure}[t]{0.32\textwidth}
        \includegraphics[width=\linewidth, height=0.625\linewidth]{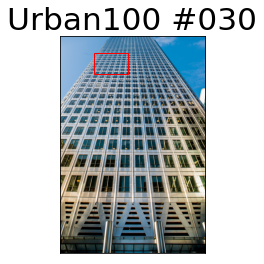} 
        \end{subfigure} 
        \hfill
        \begin{subfigure}[t]{0.32\textwidth}
        \includegraphics[width=\linewidth, height=0.625\linewidth]{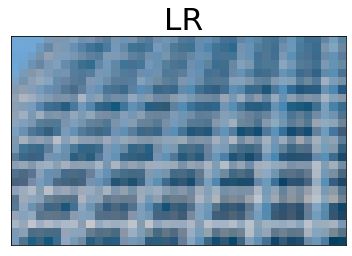}
        \end{subfigure}
        \hfill
        \begin{subfigure}[t]{0.32\textwidth}
        \includegraphics[width=\linewidth, height=0.625\linewidth]{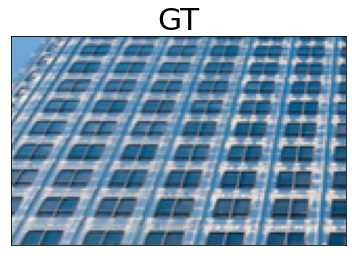}
        \end{subfigure}
    \end{subfigure}
    \begin{subfigure}[ht]{0.99\textwidth}
        \includegraphics[width=\linewidth, height=0.625\linewidth]{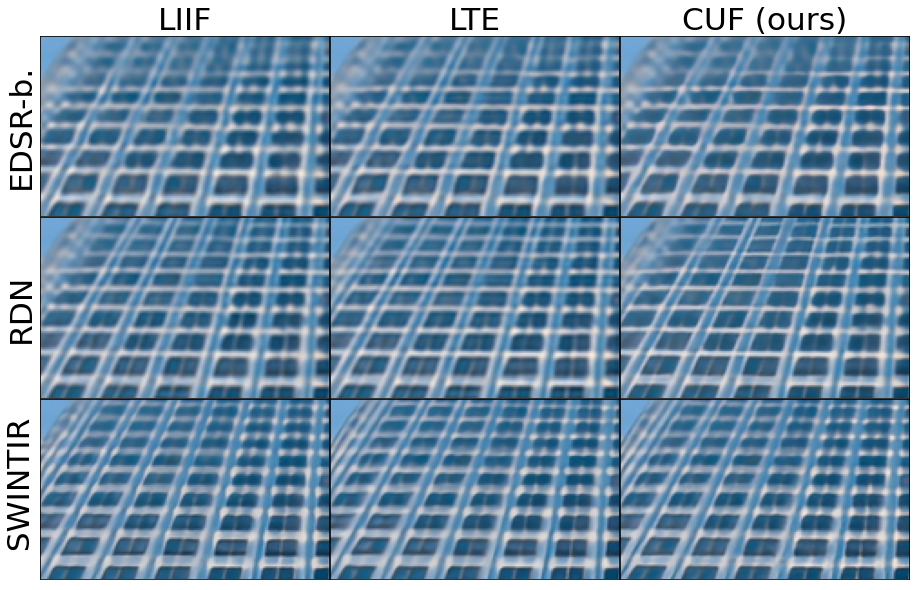}
    \end{subfigure}
\end{subfigure}
\hfill        
\begin{subfigure}{.48\linewidth}
\vfill
    \begin{subfigure}[ht]{0.99\textwidth}
        \vfill
        \begin{subfigure}[t]{0.32\textwidth}
        \includegraphics[width=\linewidth, height=0.625\linewidth]{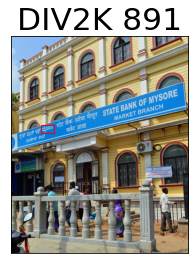} 
        \end{subfigure} 
        \hfill
        \begin{subfigure}[t]{0.32\textwidth}
        \includegraphics[width=\linewidth, height=0.625\linewidth]{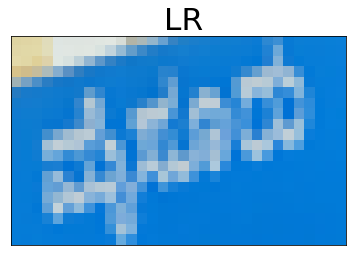}
        \end{subfigure}
        \hfill
        \begin{subfigure}[t]{0.32\textwidth}
        \includegraphics[width=\linewidth, height=0.625\linewidth]{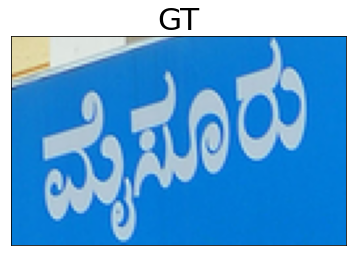}
        \end{subfigure}
    \end{subfigure}
    \begin{subfigure}[ht]{0.99\textwidth}
        \includegraphics[width=\linewidth, height=0.625\linewidth]{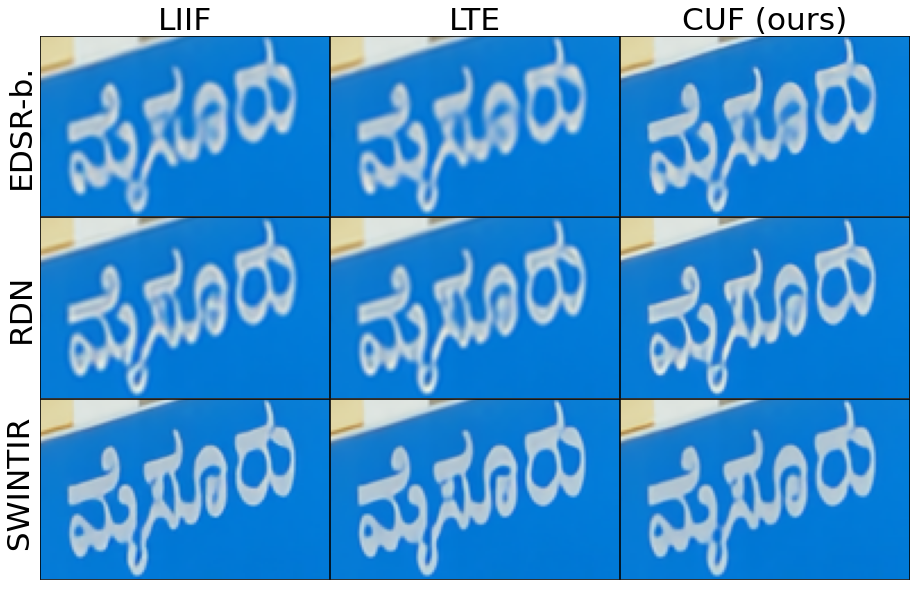}
    \end{subfigure}
\end{subfigure}
\hfill
\begin{subfigure}{.48\linewidth}
\vfill
    \begin{subfigure}[ht]{0.99\textwidth}
        \vfill
        \begin{subfigure}[t]{0.32\textwidth}
        \includegraphics[width=\linewidth, height=0.625\linewidth]{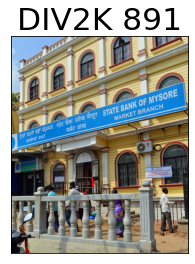} 
        \end{subfigure} 
        \hfill
        \begin{subfigure}[t]{0.32\textwidth}
        \includegraphics[width=\linewidth, height=0.625\linewidth]{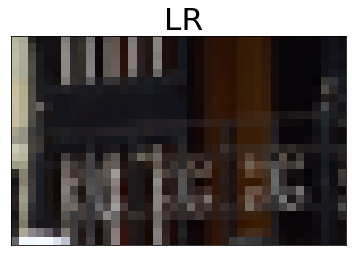}
        \end{subfigure}
        \hfill
        \begin{subfigure}[t]{0.32\textwidth}
        \includegraphics[width=\linewidth, height=0.625\linewidth]{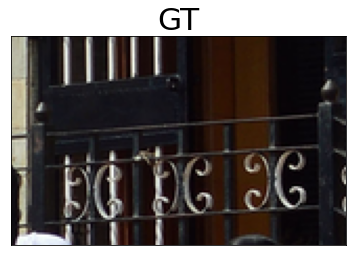}
        \end{subfigure}
    \end{subfigure}
    \begin{subfigure}[ht]{0.99\textwidth}
        \includegraphics[width=\linewidth, height=0.625\linewidth]{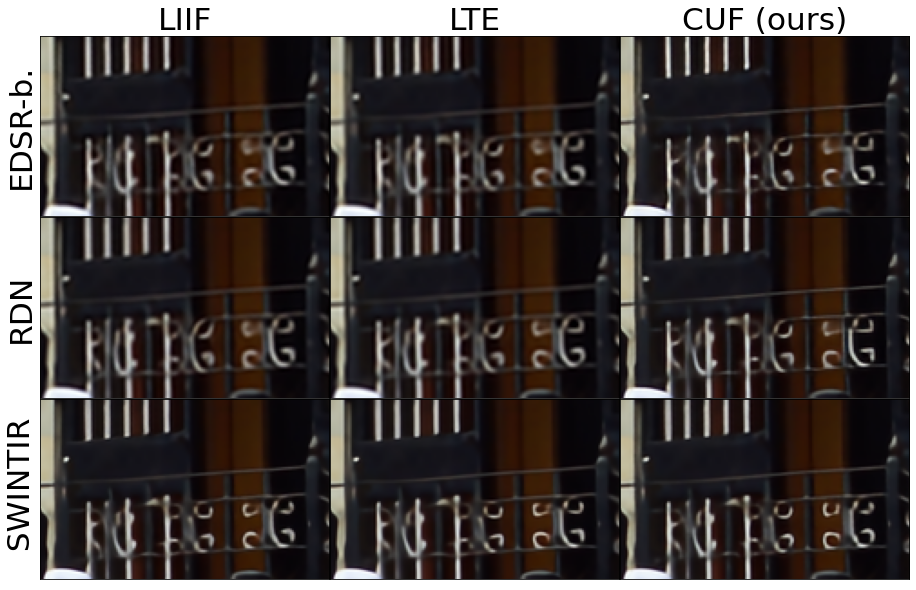}
    \end{subfigure}
\end{subfigure}
        
\caption{Qualitative comparisons of arbitrary-scale super resolution methods using different encoders. Scale factor $4\times$}
\label{fig:qual_1}
\end{figure*}

\begin{figure*}[ht!]
\centering
\begin{subfigure}{.48\linewidth}
\vfill
    \begin{subfigure}[ht]{0.99\textwidth}
        \vfill
        \begin{subfigure}[t]{0.32\textwidth}
        \includegraphics[width=\linewidth, height=0.625\linewidth]{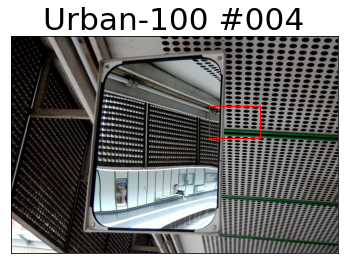} 
        \end{subfigure} 
        \hfill
        \begin{subfigure}[t]{0.32\textwidth}
        \includegraphics[width=\linewidth, height=0.625\linewidth]{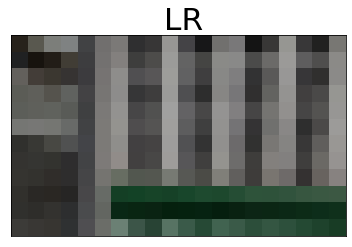}
        \end{subfigure}
        \hfill
        \begin{subfigure}[t]{0.32\textwidth}
        \includegraphics[width=\linewidth, height=0.625\linewidth]{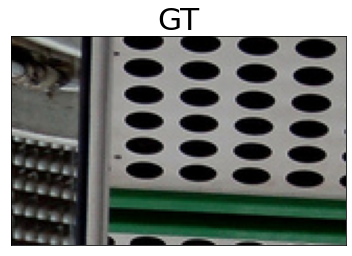}
        \end{subfigure}
    \end{subfigure}
    \begin{subfigure}[ht]{0.99\textwidth}
        \includegraphics[width=\linewidth, height=0.625\linewidth]{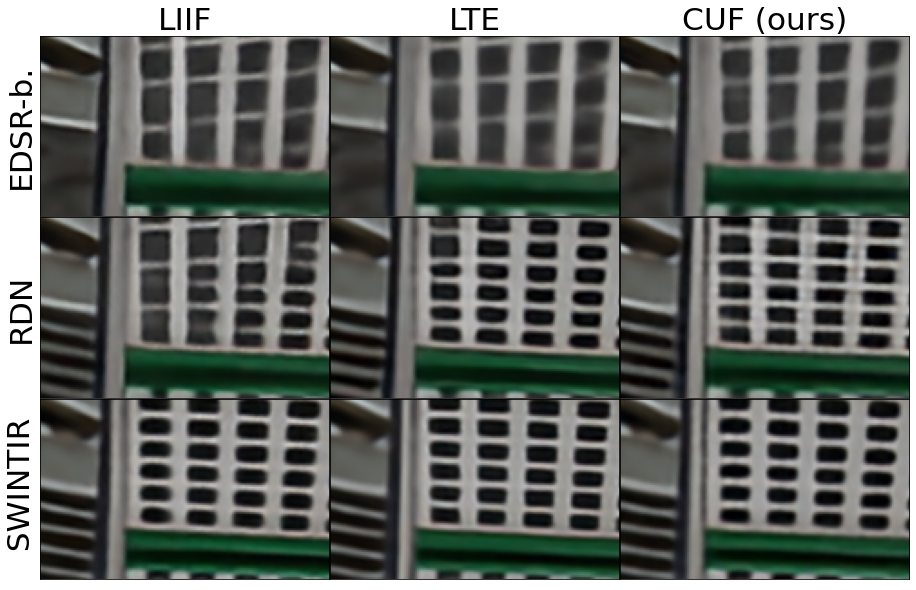}
    \end{subfigure}
\end{subfigure} 
\begin{subfigure}{.48\linewidth}
\vfill
    \begin{subfigure}[ht]{0.99\textwidth}
        \vfill
        \begin{subfigure}[t]{0.32\textwidth}
        \includegraphics[width=\linewidth, height=0.625\linewidth]{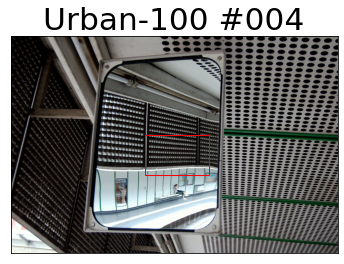} 
        \end{subfigure} 
        \hfill
        \begin{subfigure}[t]{0.32\textwidth}
        \includegraphics[width=\linewidth, height=0.625\linewidth]{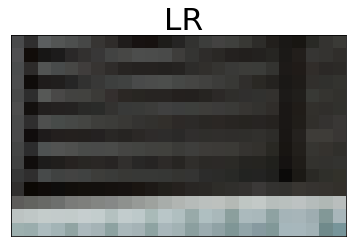}
        \end{subfigure}
        \hfill
        \begin{subfigure}[t]{0.32\textwidth}
        \includegraphics[width=\linewidth, height=0.625\linewidth]{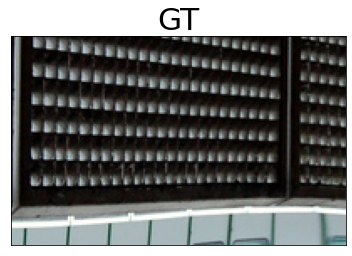}
        \end{subfigure}
    \end{subfigure}
    \begin{subfigure}[ht]{0.99\textwidth}
        \includegraphics[width=\linewidth, height=0.625\linewidth]{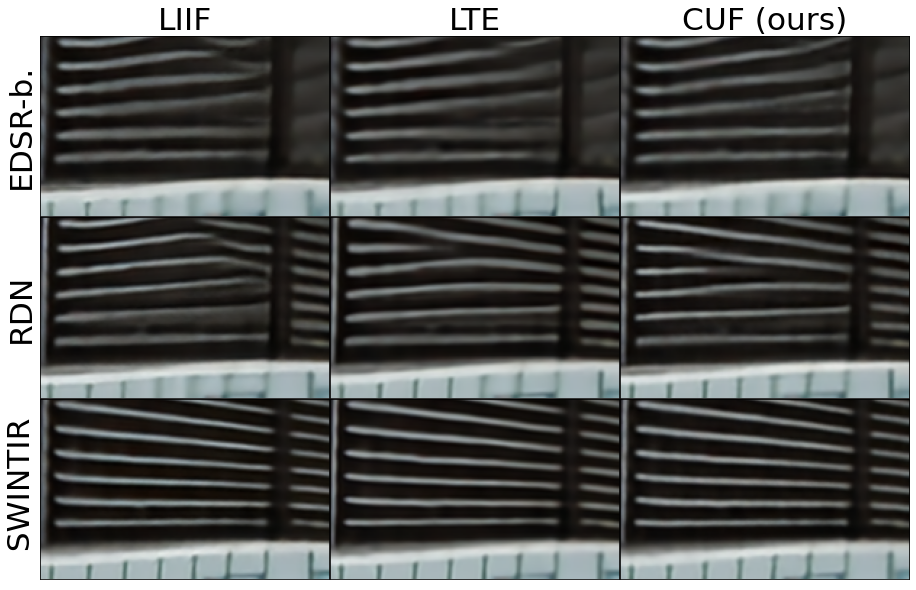}
    \end{subfigure}
\end{subfigure}
\caption{Hard cases: Aliasing artifacts observed on hard cases. Our upsampler produces the sharper results. Scale factor $8\times$}
\label{fig:qual_2}
\end{figure*}

%% file: fig/fractional_scale.tex
\begin{figure*}[ht!]
\centering
\begin{subfigure}{\textwidth}
\vfill
    \begin{subfigure}[ht]{0.03\textwidth}
        \includegraphics[width=\textwidth]{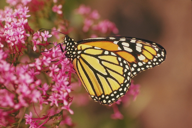}
        \caption*{LR}
    \end{subfigure}
    \begin{subfigure}[ht]{0.05\textwidth}
        \includegraphics[width=\textwidth]{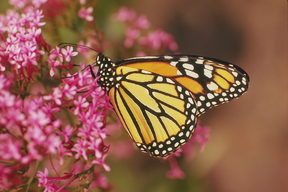}
        \caption*{$\times$1.5}
    \end{subfigure}
        \begin{subfigure}[ht]{0.07\textwidth}
        \includegraphics[width=\textwidth]{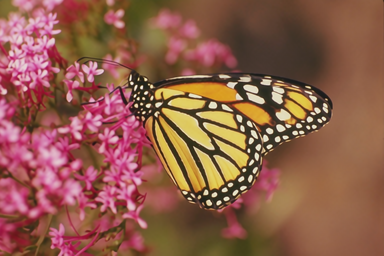}
        \caption*{$\times$2}
    \end{subfigure}
        \begin{subfigure}[ht]{0.09\textwidth}
        \includegraphics[width=\textwidth]{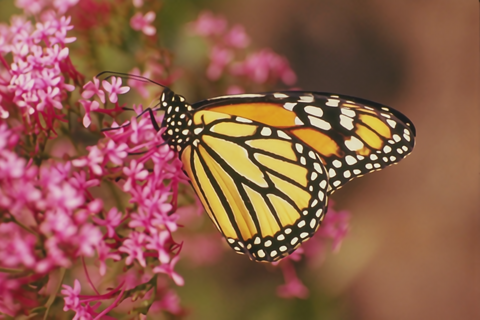}
        \caption*{$\times$2.5}
    \end{subfigure}
        \begin{subfigure}[ht]{0.11\textwidth}
        \includegraphics[width=\textwidth]{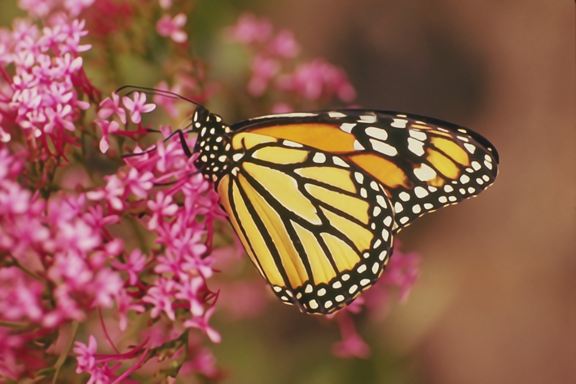}
        \caption*{$\times$3}
    \end{subfigure}
    \begin{subfigure}[ht]{0.13\textwidth}
        \includegraphics[width=\textwidth]{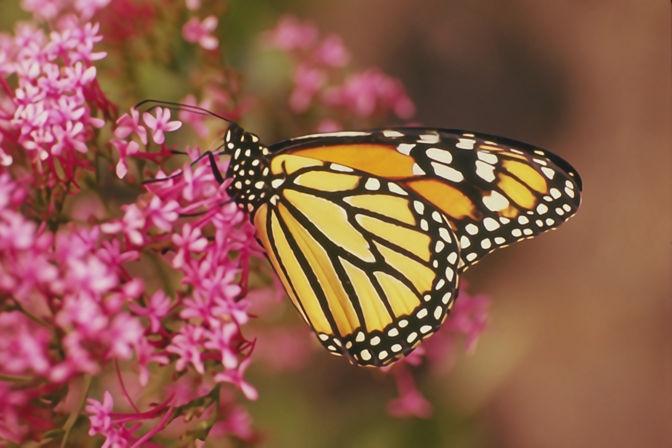}
        \caption*{$\times$3.5}
    \end{subfigure}
        \begin{subfigure}[ht]{0.14\textwidth}
        \includegraphics[width=\textwidth]{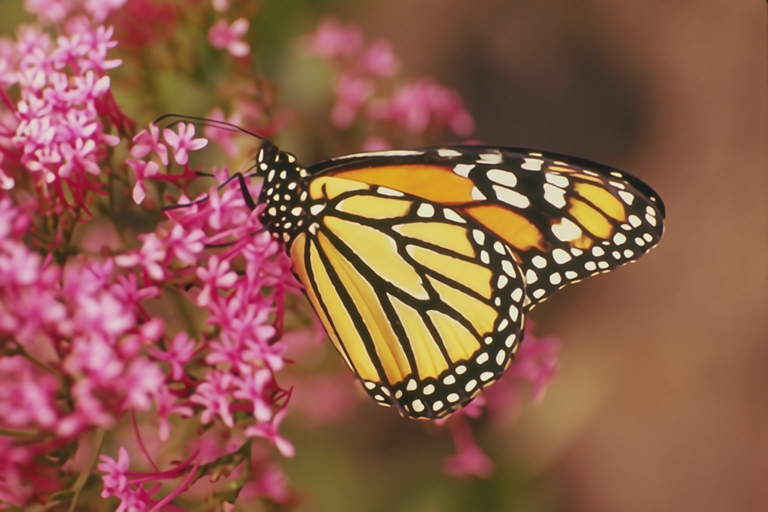}
        \caption*{$\times$4}
    \end{subfigure}
    \begin{subfigure}[ht]{0.16\textwidth}
        \includegraphics[width=\textwidth]{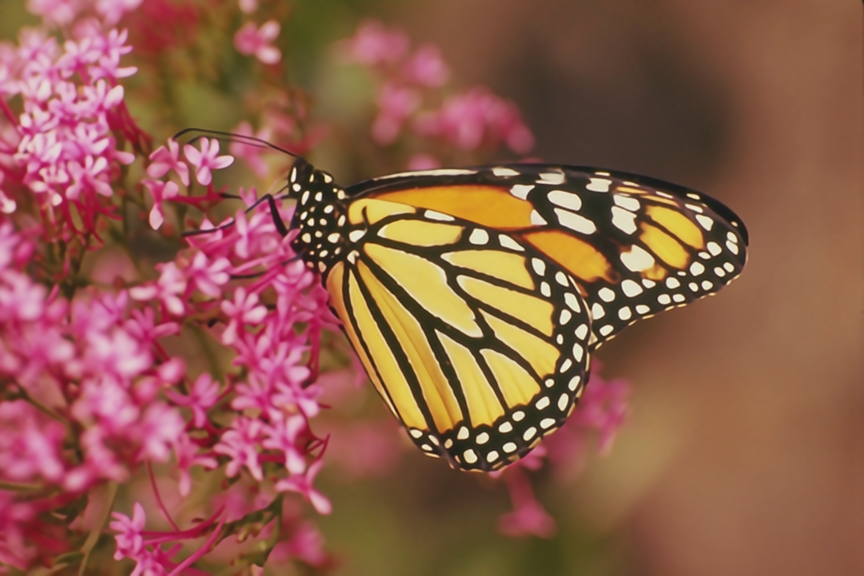}
        \caption*{$\times$4.5}
    \end{subfigure}
        \begin{subfigure}[ht]{0.18\textwidth}
        \includegraphics[width=\textwidth]{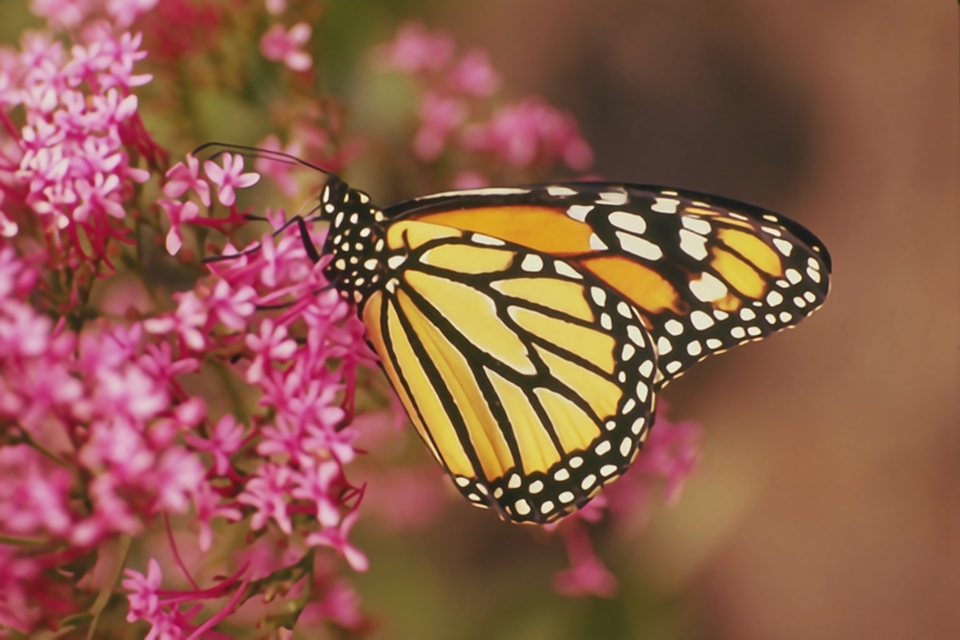}
        \caption*{$\times$5}
    \end{subfigure}
\end{subfigure}

\begin{subfigure}{\textwidth}
\vfill
    \begin{subfigure}[ht]{0.03\textwidth}
        \includegraphics[width=\textwidth]{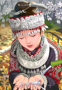}
        \caption*{LR}
    \end{subfigure}
    \begin{subfigure}[ht]{0.05\textwidth}
        \includegraphics[width=\textwidth]{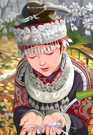}
        \caption*{$\times$1.5}
    \end{subfigure}
        \begin{subfigure}[ht]{0.07\textwidth}
        \includegraphics[width=\textwidth]{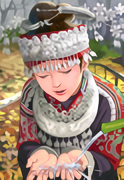}
        \caption*{$\times$2}
    \end{subfigure}
        \begin{subfigure}[ht]{0.09\textwidth}
        \includegraphics[width=\textwidth]{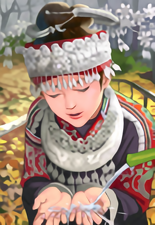}
        \caption*{$\times$2.5}
    \end{subfigure}
        \begin{subfigure}[ht]{0.11\textwidth}
        \includegraphics[width=\textwidth]{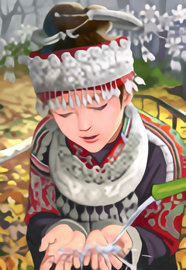}
        \caption*{$\times$3}
    \end{subfigure}
    \begin{subfigure}[ht]{0.13\textwidth}
        \includegraphics[width=\textwidth]{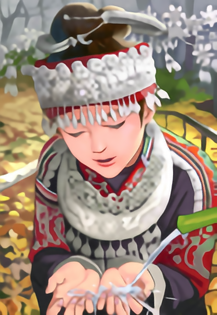}
        \caption*{$\times$3.5}
    \end{subfigure}
        \begin{subfigure}[ht]{0.14\textwidth}
        \includegraphics[width=\textwidth]{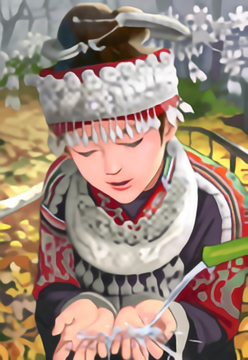}
        \caption*{$\times$4}
    \end{subfigure}
    \begin{subfigure}[ht]{0.16\textwidth}
        \includegraphics[width=\textwidth]{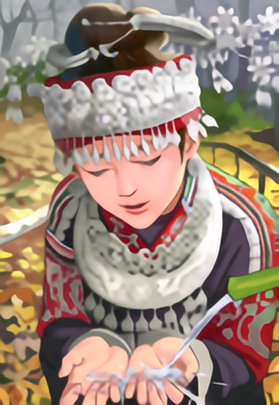}
        \caption*{$\times$4.5}
    \end{subfigure}
        \begin{subfigure}[ht]{0.18\textwidth}
        \includegraphics[width=\textwidth]{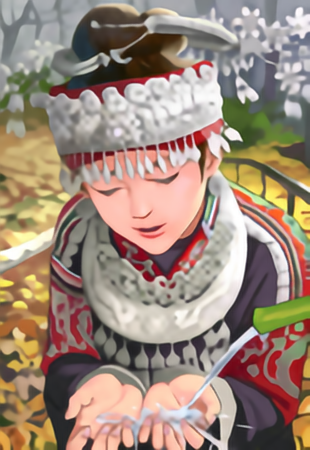}
        \caption*{$\times$5}
    \end{subfigure}
\end{subfigure}

\begin{subfigure}{\textwidth}
\vfill
    \begin{subfigure}[ht]{0.03\textwidth}
        \includegraphics[width=\textwidth]{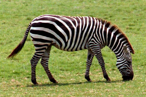}
        \caption*{LR}
    \end{subfigure}
    \begin{subfigure}[ht]{0.05\textwidth}
        \includegraphics[width=\textwidth]{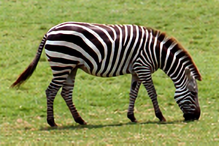}
        \caption*{$\times$1.5}
    \end{subfigure}
        \begin{subfigure}[ht]{0.07\textwidth}
        \includegraphics[width=\textwidth]{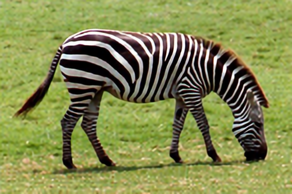}
        \caption*{$\times$2}
    \end{subfigure}
        \begin{subfigure}[ht]{0.09\textwidth}
        \includegraphics[width=\textwidth]{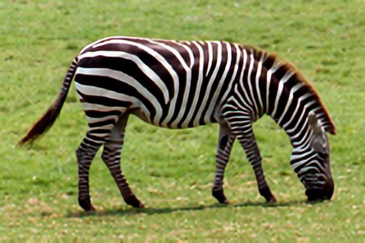}
        \caption*{$\times$2.5}
    \end{subfigure}
        \begin{subfigure}[ht]{0.11\textwidth}
        \includegraphics[width=\textwidth]{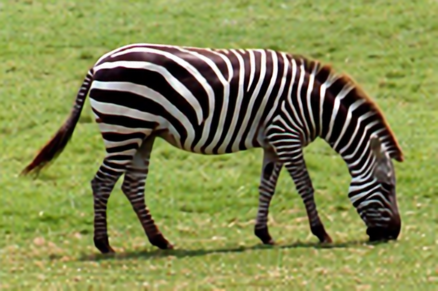}
        \caption*{$\times$3}
    \end{subfigure}
    \begin{subfigure}[ht]{0.13\textwidth}
        \includegraphics[width=\textwidth]{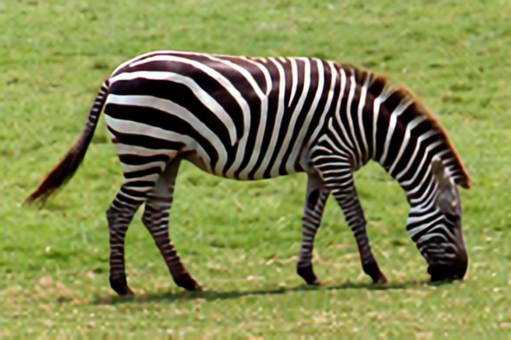}
        \caption*{$\times$3.5}
    \end{subfigure}
        \begin{subfigure}[ht]{0.14\textwidth}
        \includegraphics[width=\textwidth]{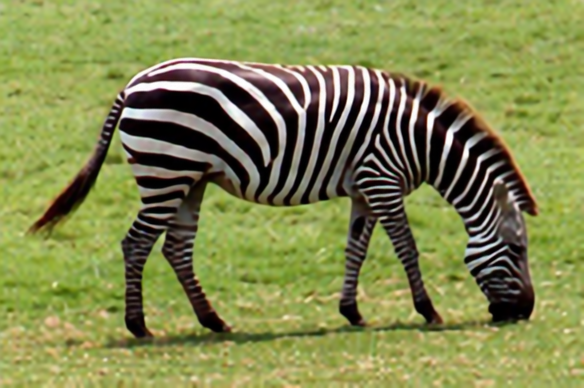}
        \caption*{$\times$4}
    \end{subfigure}
    \begin{subfigure}[ht]{0.16\textwidth}
        \includegraphics[width=\textwidth]{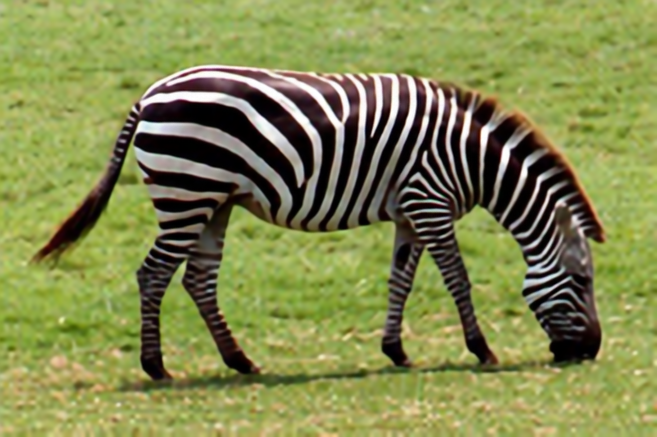}
        \caption*{$\times$4.5}
    \end{subfigure}
        \begin{subfigure}[ht]{0.18\textwidth}
        \includegraphics[width=\textwidth]{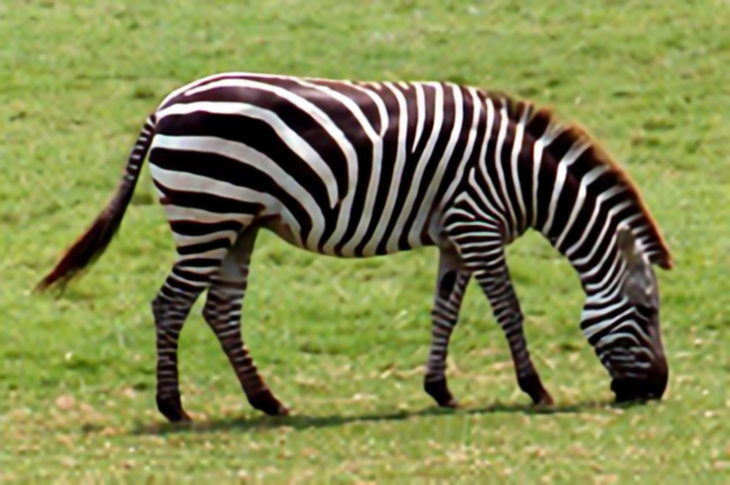}
        \caption*{$\times$5}
    \end{subfigure}
\end{subfigure}
\caption{Qualitative results using non-integer scales. Images from Set14 dataset.}
\label{fig:qual_fractional}
\end{figure*}